\documentclass{article}

\usepackage[eandd, final]{neurips_2026}

\usepackage[utf8]{inputenc}
\usepackage[T1]{fontenc}
\usepackage{hyperref}
\usepackage{url}
\usepackage{booktabs}
\usepackage{amsfonts}
\usepackage{amsmath}
\usepackage{amssymb}
\usepackage{nicefrac}
\usepackage{microtype}
\usepackage{xcolor}
\usepackage{graphicx}
\usepackage{tikz}
\usetikzlibrary{positioning, arrows.meta, calc}

\definecolor{loadred}{RGB}{220, 80, 60}
\definecolor{nodeblue}{RGB}{60, 100, 180}
\definecolor{edgegray}{RGB}{120, 120, 115}
\definecolor{encteal}{RGB}{20, 140, 110}
\definecolor{procpurp}{RGB}{100, 70, 170}
\definecolor{edgeorg}{RGB}{200, 100, 40}
\definecolor{primalgrn}{RGB}{60, 140, 50}
\definecolor{dualamb}{RGB}{190, 130, 30}
\definecolor{barrpink}{RGB}{180, 70, 120}
\definecolor{outputolv}{RGB}{140, 140, 40}
\definecolor{ipoptgray}{RGB}{90, 90, 90}
\definecolor{skipgray}{RGB}{160, 160, 155}
\definecolor{skipgreen}{RGB}{20, 120, 90}
\definecolor{annotgray}{RGB}{100, 100, 100}
\definecolor{panelbg}{RGB}{245, 244, 240}

\title{WARP: A Benchmark for Primal-Dual Warm-Starting of Interior-Point Solvers}

\author{%
  Dhruv Suri\thanks{Equal contribution.}, \hspace{0.3em}
  Helgi Hilmarsson\footnotemark[1], \hspace{0.3em} and
  Shourya Bose\footnotemark[1] \\[4pt]
  Pravah\\
  \texttt{\{dhruv.suri, helgi.hilmarsson, shourya.bose\}@pravah.com} \\
}

\begin{document}

\maketitle

\begin{abstract}
Solving AC Optimal Power Flow (AC-OPF) is of central importance in electricity
market operations, where interior-point methods (IPMs) such as IPOPT are the
standard solvers. A growing body of work uses machine learning to predict primal
warm-start iterates, reporting iteration reductions of 30--46\%. We show that
these reported gains rest on an inappropriate evaluation baseline: prior methods
benchmark against the flat start $V_m = 1, V_a = 0$, whereas the solver's actual
default---the variable-bound midpoint $(l+u)/2$---is near-optimal for log-barrier
centrality. Against this corrected baseline, no primal-only warm-start method
reduces solver iterations. We trace the failure to a geometric property of
interior-point methods: primal prediction accuracy is anticorrelated with
convergence speed, and providing the ground-truth optimal solution $x^*$ without
dual variables causes the solver to diverge. Oracle experiments establish that
the complete primal-dual-barrier state $(x^*, \lambda^*, z^*, \mu^*)$ reduces
IPOPT iterations from 23 to 3---an 85\% reduction that is structurally
inaccessible to primal-only methods. To enable rigorous evaluation of warm-start
methods on this task, we release a benchmark suite comprising dual-labeled AC-OPF
datasets with IPOPT-extracted solutions, a corrected evaluation protocol, and
WARP---a topology-conditioned encode-process-decode interaction network that
predicts the full interior-point state $(\hat{x}, \hat{\lambda}, \hat{z},
\hat{\mu})$ on the heterogeneous constraint graph. WARP achieves a 76\%
reduction in IPOPT iterations while natively accommodating N-1 contingency
topology variations without retraining.
\end{abstract}

\section{Introduction}
\label{sec:intro}

Interior-point methods (IPMs) are the dominant solvers for large-scale constrained nonlinear programs~\citep{nocedal1999numerical, nesterov1994interior}. In power systems, the primal-dual IPM solver IPOPT~\citep{wachter2006ipopt} solves the AC Optimal Power Flow (AC-OPF) problem---a non-convex NLP determining minimum-cost generator dispatch subject to Kirchhoff's laws and operational limits~\citep{cain2012acopf}---every 5--15 minutes at independent system operators worldwide. Its computational cost motivates a substantial body of work on learning-based acceleration.

The prevailing approach trains a neural network to predict primal warm-start iterates $\hat{x}$, reporting 30--46\% iteration reductions over the flat start~\citep{baker2019learning, diehl2019warm, zhang2022fast, cao2023fast, xu2025explainable}. These results suggest a straightforward principle: predict $\hat{x}$ closer to $x^*$, and the solver converges faster. We demonstrate that this principle is structurally misaligned with how interior-point methods converge, and that the reported gains rest on an inappropriate baseline.

Through systematic experimentation, we establish three findings:
\begin{enumerate}
    \item \textbf{The evaluation baseline is inappropriate.} Prior methods benchmark against the flat start $V_m = 1, V_a = 0$---not the solver's default. IPOPT and MIPS initialise at the variable-bound midpoint $(l+u)/2$, which is near-optimal for log-barrier centrality. Against this corrected baseline, no primal-only method reduces solver iterations.
    \item \textbf{Primal accuracy is anticorrelated with convergence.} Among model variants, lower prediction error yields \emph{more} solver iterations. Providing the ground-truth $x^*$ without duals causes divergence, confirming the theoretical result of~\citet{yildirim2002warm}: the optimal solution lies at constraint boundaries where the log-barrier is singular.
    \item \textbf{Dual information is necessary and sufficient.} Oracle experiments show that the complete state $(x^*, \lambda^*, z^*, \mu^*)$ reduces iterations by 85\%---a ceiling structurally inaccessible to primal-only methods. The entire iteration gap resides in the dual variables and barrier parameter.
\end{enumerate}

\paragraph{Contributions.} We make three contributions:
\begin{enumerate}
    \item \textbf{A corrected evaluation protocol and dual-labeled dataset.} We release AC-OPF datasets with IPOPT-extracted primal-dual-barrier solutions $(x^*, \lambda^*, z^*, \mu^*)$, hosted on OpenML with Croissant metadata. The evaluation protocol benchmarks against the solver's true midpoint default rather than the weak flat start employed in prior work.\footnote{We also document a silent initialisation defect in pandapower's PIPS solver that overwrites user-supplied warm-starts; see Appendix~\ref{app:pips_bug}.}
    \item \textbf{A systematic diagnosis of the primal-only failure.} Fifteen ablation experiments---spanning prediction, blending, projection, retraction, and constraint screening---characterise the geometric bottleneck and establish that dual information is both necessary and sufficient for effective IPM warm-starting.
    \item \textbf{WARP: a topology-aware primal-dual baseline.} An encode-process-decode interaction network that predicts the full IPM state, achieving 76\% iteration reduction. Two architectural principles are identified: per-step edge feature updates dominate ($9\times$ impact), and omitting node residual connections improves dual prediction by $5\times$.
\end{enumerate}

\section{Background}
\label{sec:background}

\subsection{AC optimal power flow and interior-point methods}

The AC-OPF problem determines minimum-cost generator dispatch on a power grid $\mathcal{G} = (\mathcal{B}, \mathcal{E})$ with buses $\mathcal{B}$ and branches $\mathcal{E}$. Decision variables $x = (V_a, V_m, P_g, Q_g)$ comprise bus voltage angles and magnitudes and generator real and reactive power. The problem minimises a quadratic cost $\sum_g c_{2g}P_g^2 + c_{1g}P_g + c_{0g}$ subject to AC power balance $P_i^{\mathrm{inj}} + jQ_i^{\mathrm{inj}} = V_i \sum_k Y_{ik}^* V_k^*$, voltage limits, thermal line ratings, and generator bounds (full formulation in Appendix~\ref{app:acopf}). The power balance constraints are non-convex, making AC-OPF a non-convex NLP.

The standard solver is the primal-dual interior-point method~\citep{wachter2006ipopt, wang2007mips}. The IPM replaces bound constraints with a logarithmic barrier parameterised by $\mu > 0$ and applies Newton's method to the perturbed KKT system, maintaining the state $(x, \lambda, z_l, z_u, \mu)$ where $\lambda \in \mathbb{R}^m$ are equality multipliers and $z_l, z_u \in \mathbb{R}^n_+$ are bound multipliers. The complementarity conditions $(x_i - l_i)z_{l,i} = \mu$ define the \emph{central path}---a curve of well-centred iterates from large $\mu$ (deep interior) to $\mu \to 0$ (the boundary solution). The centrality measure $\bar{\mu} = ((x-l)^\top z_l + (u-x)^\top z_u) / 2n$ quantifies how well-centred an iterate is.

Both IPOPT and MIPS initialise at the variable-bound midpoint:
\begin{equation}
    x_0 = (l + u) / 2
    \label{eq:midpoint}
\end{equation}
This maximises the minimum slack $\min_i \min(x_i - l_i, u_i - x_i)$ and yields large $\bar{\mu}$. A critical consequence formalised by~\citet{yildirim2002warm} is that $x^*$ constitutes the \emph{least favorable} IPM starting point: active-constraint slacks vanish, rendering the barrier singular. IPOPT supports warm-starting via \texttt{warm\_start\_init\_point}, accepting user-supplied $(x_0, \lambda_0, z_{l,0}, z_{u,0})$ with $\mu_{\mathrm{init}}$. The configuration \texttt{warm\_start\_bound\_push}$=10^{-20}$, \texttt{mu\_strategy=monotone}, introduced by~\citet{gao2024ipmlstm}, is adopted throughout.

\subsection{Prior work}

\paragraph{Primal warm-starting.} \citet{baker2019learning} trained random forests, \citet{diehl2019warm} used GNNs, \citet{zhang2022fast} and \citet{cao2023fast} used decision trees, and \citet{xu2025explainable} proposed ensemble methods for predicting primal warm-starts. All report 30--46\% reductions \emph{relative to the flat start} $V_m = 1, V_a = 0$---not the solver's actual midpoint default~\eqref{eq:midpoint}. \citet{deihim2024initial} and \citet{liu2022topology} applied GNNs with physics regularisation. All predict only $x$; the solver reinitialises duals independently.

\paragraph{Primal-dual methods.} \citet{gao2024ipmlstm} introduced IPM-LSTM, predicting $(x, \lambda, z)$ via coordinate-wise LSTMs trained on KKT trajectories. On QPs with up to 200 variables, it achieves 63.9\% iteration reduction. However, it operates on a fixed-dimension KKT vector---necessitating a separate model per topology---and has not been applied to AC-OPF. \citet{klamkin2024dual} proposed dual completion layers; \citet{sambharya2024l2ws} learned warm-starts for fixed-point splitting; \citet{briden2024toast} targeted SQP solvers.

\paragraph{End-to-end and constraint methods.} Solution prediction methods~\citep{fioretto2020predicting, donti2021dc3, park2023self, chen2024deep} replace the solver entirely. Active constraint identification~\citep{misra2022learning, deka2019learning, pineda2020screening} reduces problem dimensionality. Both are complementary; we show constraint screening is counterproductive for IPM (Section~\ref{sec:primal_fails}).

\paragraph{GNN architectures.} WARP belongs to the encode-process-decode (EPD) family~\citep{battaglia2016interaction, battaglia2018relational} applied to physical simulation~\citep{pfaff2021learning, lam2023graphcast} and AC-OPF~\citep{piloto2024canos, rivera2025pfdelta}. These predict only primal variables; we extend to the full primal-dual-barrier state. Extended discussion in Appendix~\ref{app:related}.

\section{The Primal-Only Warm-Start Failure}
\label{sec:primal_fails}

All experiments use \texttt{pglib\_opf\_case118\_ieee} from OPFDataset~\citep{falconer2023opfdataset} (67,500 training, 3,750 validation instances; 50 held-out test instances). We evaluate on PIPS~\citep{wang2007mips} (via pandapower) and IPOPT~\citep{wachter2006ipopt} (via cyipopt with exact Hessian and sparse Jacobian).

\subsection{Correcting the baseline and systematic ablation}

Prior work evaluates against the flat start $V_m = 1, V_a = 0$, which is not the solver's default. The midpoint~\eqref{eq:midpoint} achieves 19.6 mean PIPS iterations and 22.6 mean IPOPT iterations. We additionally identified a silent defect in pandapower's PIPS implementation that overwrites user-supplied initial points (Appendix~\ref{app:pips_bug}).

We train a heterogeneous GNN (DetGNN) predicting $(V_a, V_m, P_g, Q_g)$ from load data and evaluate eight strategies for injecting predictions into the solver. Table~\ref{tab:primal_only} reports the results.

\begin{table}[h]
\caption{Primal-only warm-start strategies on case118 (PIPS, 50 test instances). Per-variable normalisation is applied below the second divider. No method improves upon the solver's midpoint default.}
\label{tab:primal_only}
\centering
\small
\begin{tabular*}{\textwidth}{@{\extracolsep{\fill}}llcc@{}}
\toprule
Method & Description & Mean iters & vs.\ Midpoint \\
\midrule
Midpoint $(l+u)/2$ & Solver default & 19.6 & --- \\
\midrule
DetGNN (unnorm.) & Primal $\hat{x}$ & 31.0 & $+58\%$ \\
WARP diffusion $K\!=\!3$ & Primal $\hat{x}$ & 31.0 & $+58\%$ \\
\midrule
DetGNN (norm.) & Primal $\hat{x}$ & 21.1 & $+7.8\%$ \\
Selective (voltage only) & Partial $\hat{x}$ & 21.6 & $+10\%$ \\
Barrier retraction & $\hat{x}$ retracted to interior & 21.3 & $+8.7\%$ \\
Feasibility projection & $\hat{x}$ projected to bounds & 30.9 & $+58\%$ \\
Blend ($\alpha\!=\!0.5$) & $\alpha\hat{x} + (1\!-\!\alpha)x_{\mathrm{mid}}$ & 25.1 & $+28\%$ \\
Constraint screening & Reduced problem & 44.1 & $+125\%$ \\
\bottomrule
\end{tabular*}
\end{table}

\textbf{No primal-only method improves upon the midpoint.} Per-variable normalisation---addressing a $100\times$ scale disparity between $P_g$ and $V_m$---was the single most impactful change, improving from 31.0 to 21.1 iterations, but still 7.8\% worse than the midpoint. Feasibility projection and barrier retraction were ineffective: projecting predictions onto the strict interior with $\epsilon = 0.02$ yields 30.9 iterations; retraction to restore centrality achieves at best 21.3. Constraint screening was catastrophically counterproductive ($44.1$ iterations, $2.3\times$ the midpoint), consistent with the observation that removing constraints widens the feasible set and flattens the barrier landscape. Full per-strategy analysis in Appendix~\ref{app:primal_ablation}.

\subsection{The centrality-accuracy tradeoff}
\label{sec:centrality}

A blend sweep $x_0 = \alpha\hat{x}_{\mathrm{GNN}} + (1-\alpha)x_{\mathrm{mid}}$ for $\alpha \in [0, 1]$ reveals a counterintuitive result: \emph{any intermediate blend is worse than either endpoint}, reaching 28.8 iterations at $\alpha = 0.1$. The midpoint ($\alpha=0$, 19.6 iters) and pure GNN prediction ($\alpha=1$, 21.6 iters) are both local minima. This indicates that the interpolation path between a well-centred interior point and an accurate near-boundary point traverses regions of high barrier curvature.

Among six diffusion model variants trained with varying physics loss weights, the variant with the lowest denoising loss ($\mathcal{L}_{\mathrm{ddpm}} = 0.265$) yields the worst solver performance (39.5 iterations), while the variant with the highest denoising loss ($\mathcal{L}_{\mathrm{ddpm}} = 0.576$) yields the best (28.8 iterations). Minimising prediction error in solution space is a misaligned objective for IPM warm-starting; physical consistency matters more than pointwise accuracy.

The geometric bottleneck is voltage magnitude $V_m$. Its feasible range on case118 is only 0.12 p.u.\ ($V_m \in [0.94, 1.06]$), so any prediction displacing $V_m$ from the midpoint (1.00) rapidly approaches constraint boundaries, degrading the log-barrier irrespective of accuracy. Selective warm-start experiments confirm that generator variables contribute negligibly to the iteration count (Appendix~\ref{app:primal_ablation}).

\subsection{Oracle experiments: the primal-dual-barrier state is necessary and sufficient}
\label{sec:oracle}

We construct a direct cyipopt interface with exact Hessian, sparse Jacobian, and the full warm-start protocol. Table~\ref{tab:oracle} reports oracle warm-start results.

\begin{table}[h]
\caption{Oracle warm-start experiments (IPOPT, case118, 50 instances). The complete primal-dual-barrier state yields $85\%$ reduction; primal variables alone contribute negatively.}
\label{tab:oracle}
\centering
\small
\begin{tabular*}{\textwidth}{@{\extracolsep{\fill}}lccc@{}}
\toprule
Warm-start content & Mean iters & Median & Reduction \\
\midrule
Cold start (midpoint) & 22.6 & 22 & --- \\
Oracle primal $x^*$ only & 23.7 & 23 & $-4.9\%$ \\
Ground-truth $x^*$ (cold dual init) & $>$50 & --- & Diverges \\
\midrule
Oracle $(x^*, \lambda^*)$ & 12.4 & 12 & $+45\%$ \\
Oracle $(x^*, \lambda^*, z^*)$ & 4.7 & 5 & $+79\%$ \\
Oracle $(x^*, \lambda^*, z^*, \mu^*)$ & \textbf{3.3} & \textbf{3} & $\mathbf{+85\%}$ \\
\bottomrule
\end{tabular*}
\end{table}

Three results merit emphasis. First, the oracle primal $x^*$ \emph{without} duals yields 23.7 iterations---\emph{worse} than cold start (22.6), confirming that primal accuracy alone does not accelerate IPM convergence. Second, providing $x^*$ with cold dual initialisation causes divergence ($>$50 iterations), experimentally confirming~\citet{yildirim2002warm}. Third, the complete state achieves 3.3 iterations (85\% reduction), establishing the ceiling.

The decomposition is instructive: bound multipliers $z^*$ contribute substantially more than equality multipliers $\lambda^*$ alone ($12.4 \to 4.7$); the barrier parameter $\mu^*$ adds a further reduction to 3.3. The primal variables contribute \emph{negatively} in isolation. The entire iteration gap between cold start and oracle resides in the dual and barrier state.

\section{The WARP Benchmark Suite}
\label{sec:benchmark}

The findings of Section~\ref{sec:primal_fails} motivate a benchmark that evaluates warm-start methods on the correct objective: predicting the full primal-dual-barrier state $(x, \lambda, z, \mu)$ and measuring the resulting iteration count against the solver's true midpoint default.

\subsection{Dual-labeled datasets}
\label{sec:datasets}

Existing AC-OPF datasets~\citep{falconer2023opfdataset, rivera2025pfdelta} provide only primal solution labels. We augment each instance with the complete IPOPT-converged state:
\begin{equation}
    \mathcal{D}_i = \left(x_i^*,\; \lambda_i^*,\; z_{l,i}^*,\; z_{u,i}^*,\; \mu_i^*,\; f(x_i^*)\right)
    \label{eq:dual_label}
\end{equation}
For \texttt{pglib\_opf\_case118\_ieee}: 5,000 training, 500 validation, and 50 test instances, with 100\% IPOPT convergence at $\sim$2.5 seconds per instance on a single CPU core. The dataset is hosted on OpenML with Croissant metadata (Appendix~\ref{app:datasheet}).

The dual variables present distinct challenges for regression. Equality multipliers $\lambda^*$ span $[-14, 4168]$ with std 1,564. Bound multipliers $z_l^*, z_u^*$ are 72--85\% sparse (exactly zero for non-binding constraints) with heavy-tailed non-zero entries. The barrier parameter $\mu^* \approx 3.25 \times 10^{-8}$ is approximately constant across instances. Per-variable normalisation to $\mathcal{N}(0,1)$ was the critical enabler: without it, the MSE loss is dominated by high-magnitude $\lambda^*$, and the model fails to learn the small-but-critical bound multipliers. Normalisation reduced dual prediction error from $9.45 \times 10^5$ to $3 \times 10^{-4}$ (normalised MSE).

\subsection{Evaluation protocol}
\label{sec:protocol}

We define a standardised evaluation protocol: (1)~\textbf{Solver}: IPOPT via cyipopt, exact Hessian, sparse Jacobian, tolerance $10^{-4}$. (2)~\textbf{Warm-start configuration}: \texttt{warm\_start\_init\_point=yes}, \texttt{bound\_push}$=10^{-20}$, \texttt{mu\_strategy=monotone}, \texttt{mu\_init}$=\hat{\mu}$. (3)~\textbf{Baseline}: the solver's midpoint default $(l+u)/2$---\emph{not} the flat start. (4)~\textbf{Metric}: IPOPT iteration count via \texttt{intermediate()} callback; mean, median, and per-instance counts. (5)~\textbf{Oracle ceiling}: ground-truth $(x^*, \lambda^*, z^*, \mu^*)$.

This protocol addresses the two methodological deficiencies identified in Section~\ref{sec:primal_fails}: the inappropriate flat-start baseline and the absence of dual variable evaluation.

\subsection{Metrics}

We evaluate along three axes. \textbf{Solver convergence}: the primary metric is the IPOPT iteration count. \textbf{Prediction quality}: normalised MSE decomposed into primal and dual components, and per-variable Pearson correlation. The decomposition is essential---a model with low total loss but poor dual prediction yields high iteration counts regardless of primal quality (Section~\ref{sec:oracle}). \textbf{Topology robustness}: N-1 contingency scenarios (single line removals, no retraining). For topology-blind models, we report whether they can process the modified topology at all.

\section{WARP: Topology-Aware Primal-Dual Prediction}
\label{sec:method}

The oracle results establish that predicting the full state $(x, \lambda, z, \mu)$ is both necessary and sufficient. WARP performs this prediction while operating directly on the grid topology.

\subsection{Problem formulation}

Given grid $\mathcal{G} = (\mathcal{B}, \mathcal{E})$ with loads $(P^d, Q^d)$, the task is:
\begin{equation}
    f_\theta: (\mathcal{G}, P^d, Q^d) \mapsto (\hat{x}, \hat{\lambda}, \hat{z}_l, \hat{z}_u, \hat{\mu})
    \label{eq:warp_map}
\end{equation}
The model must satisfy two requirements beyond pointwise accuracy: (i)~the predicted state should be well-centred (complementarity products approximately uniform), and (ii)~it should operate on the graph structure of $\mathcal{G}$, enabling application to varying topologies without retraining.

\subsection{Architecture}
\label{sec:architecture}

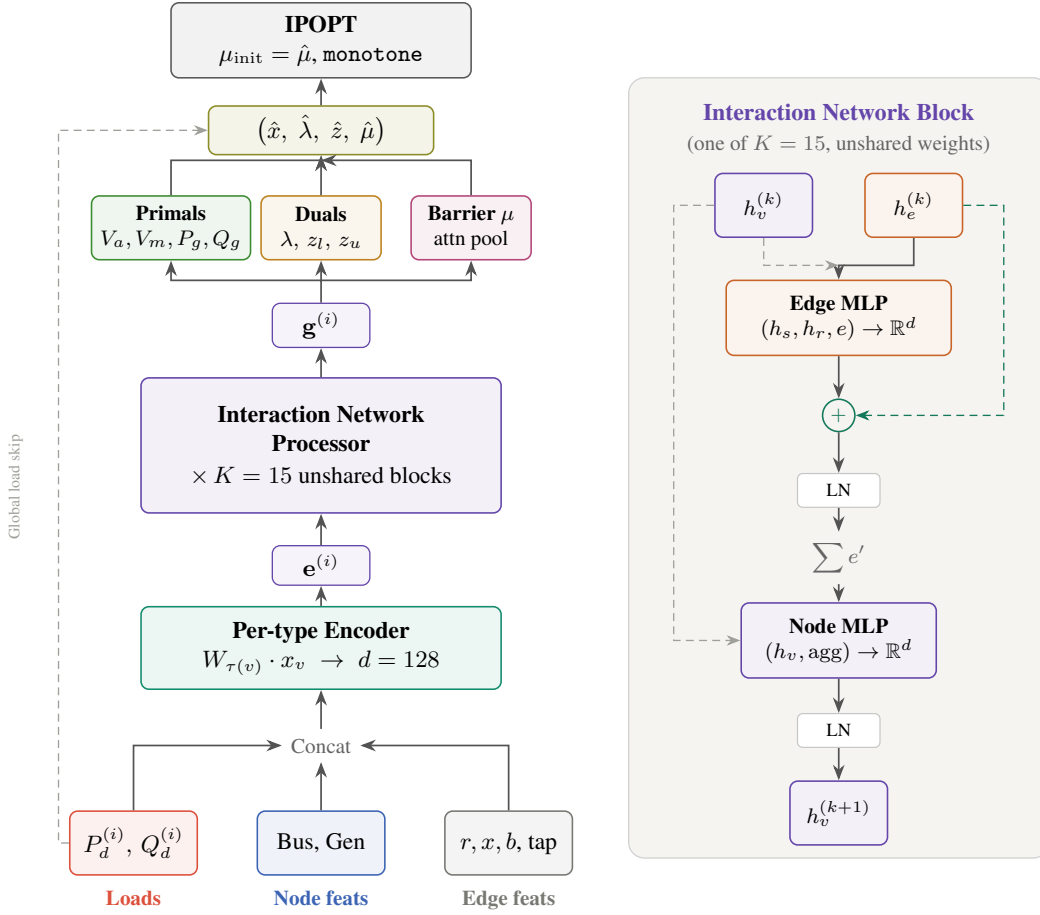
\begin{figure}[htbp!]
\centering
\resizebox{\textwidth}{!}{%
\begin{tikzpicture}[
    >=Stealth,
    every node/.style={font=\small},
    box/.style 2 args={draw=#1, fill=#1!8, rounded corners=3pt,
        minimum height=0.85cm, minimum width=#2, line width=0.6pt, align=center},
    statebox/.style={draw=procpurp, fill=procpurp!6, rounded corners=3pt,
        minimum height=0.5cm, minimum width=1.3cm, line width=0.5pt},
    arr/.style={->, line width=0.65pt, color=gray!65!black},
    skiparr/.style={->, densely dashed, line width=0.5pt, color=skipgray},
]
 
 
\node[box={loadred}{1.7cm}] (load) at (0, 0) {$P_d^{(i)},\, Q_d^{(i)}$};
\node[font=\footnotesize\bfseries, color=loadred, below=2pt of load] {Loads};
 
\node[box={nodeblue}{1.7cm}] (nfeat) at (2.5, 0) {Bus, Gen};
\node[font=\footnotesize\bfseries, color=nodeblue, below=2pt of nfeat] {Node feats};
 
\node[box={edgegray}{1.7cm}] (efeat) at (5.0, 0) {$r, x, b$, tap};
\node[font=\footnotesize\bfseries, color=edgegray, below=2pt of efeat] {Edge feats};
 
\node[font=\footnotesize, color=annotgray] (concat) at (2.5, 1.3) {Concat};
\draw[arr] (load.north) -- (0, 1.3) -- (concat);
\draw[arr] (nfeat.north) -- (concat);
\draw[arr] (efeat.north) -- (5.0, 1.3) -- (concat);
 
\node[box={encteal}{4.8cm}, minimum height=1.1cm] (encoder) at (2.5, 2.6) {%
    \textbf{Per-type Encoder}\\[1pt]
    $W_{\tau(v)} \cdot x_v \;\to\; d = 128$};
\draw[arr] (concat) -- (encoder);
 
\node[statebox] (latent_e) at (2.5, 3.7) {$\mathbf{e}^{(i)}$};
\draw[arr] (encoder) -- (latent_e);
 
\node[box={procpurp}{4.8cm}, minimum height=1.8cm] (processor) at (2.5, 5.3) {%
    \textbf{Interaction Network}\\[1pt]
    \textbf{Processor}\\[2pt]
    $\times\, K = 15$ unshared blocks};
\draw[arr] (latent_e) -- (processor);
 
\node[statebox] (latent_g) at (2.5, 6.9) {$\mathbf{g}^{(i)}$};
\draw[arr] (processor) -- (latent_g);
 
\node[box={primalgrn}{2.0cm}, font=\footnotesize] (dec_p) at (0.5, 8.2) {%
    \textbf{Primals}\\$V_a, V_m, P_g, Q_g$};
\node[box={dualamb}{1.6cm}, font=\footnotesize] (dec_d) at (2.5, 8.2) {%
    \textbf{Duals}\\$\lambda,\, z_l,\, z_u$};
\node[box={barrpink}{1.6cm}, font=\footnotesize] (dec_m) at (4.5, 8.2) {%
    \textbf{Barrier} $\mu$\\attn pool};
 
\draw[arr] (latent_g) -- (2.5, 7.5) -- (0.5, 7.5) -- (dec_p);
\draw[arr] (2.5, 7.5) -- (dec_d);
\draw[arr] (2.5, 7.5) -- (4.5, 7.5) -- (dec_m);
 
\node[draw=outputolv, fill=outputolv!10, rounded corners=3pt,
    minimum height=0.55cm, minimum width=3.0cm, line width=0.6pt]
    (output) at (2.5, 9.5) {$\bigl(\hat{x},\; \hat{\lambda},\; \hat{z},\; \hat{\mu}\bigr)$};
 
\draw[arr] (dec_p.north) -- (0.5, 9.1) -- (2.5, 9.1) -- (output);
\draw[arr] (dec_d) -- (output);
\draw[arr] (dec_m.north) -- (4.5, 9.1) -- (2.5, 9.1);
 
\node[box={ipoptgray}{4.0cm}, minimum height=1.0cm] (ipopt) at (2.5, 10.7) {%
    \textbf{IPOPT}\\[1pt]
    $\mu_{\mathrm{init}} = \hat{\mu}$, \texttt{monotone}};
\draw[arr] (output) -- (ipopt);
 
\draw[skiparr]
    (load.west) -- (-1.0, 0)
    -- (-1.0, 9.5)
    -- (output.west);
\node[font=\tiny, color=skipgray, rotate=90, anchor=south] at (-1.35, 4.75) {Global load skip};
 
\begin{scope}[xshift=9.2cm, yshift=0.0cm]
 
 
\fill[panelbg, rounded corners=6pt] (-2.6, -0.2) rectangle (3.0, 10.2);
\draw[gray!35, rounded corners=6pt, line width=0.5pt] (-2.6, -0.2) rectangle (3.0, 10.2);
 
\node[font=\small\bfseries, color=procpurp] at (0.2, 9.75) {Interaction Network Block};
\node[font=\footnotesize, color=annotgray] at (0.2, 9.3) {(one of $K = 15$, unshared weights)};
 
\node[box={procpurp}{1.3cm}, font=\footnotesize] (hv_in) at (-0.8, 8.5) {$h_v^{(k)}$};
\node[box={edgeorg}{1.3cm}, font=\footnotesize] (he_in) at (1.2, 8.5) {$h_e^{(k)}$};
 
\node[box={edgeorg}{3.0cm}, minimum height=1.0cm, font=\footnotesize]
    (edge_mlp) at (0.2, 7.0) {\textbf{Edge MLP}\\$(h_s, h_r, e) \to \mathbb{R}^d$};
 
\draw[arr] (he_in.south) -- (1.2, 7.7) -- (0.2, 7.7) -- (edge_mlp.north);
\draw[skiparr] (hv_in.south) -- (-0.8, 7.7) -- (0.2, 7.7);
 
\node[draw=skipgreen, circle, inner sep=1.5pt, font=\scriptsize\bfseries,
    text=skipgreen, line width=0.6pt] (eplus) at (0.2, 5.7) {$+$};
\draw[arr] (edge_mlp) -- (eplus);
 
\draw[densely dashed, skipgreen, line width=0.5pt, ->]
    (he_in.east) -- (2.4, 8.5)
    -- (2.4, 5.7)
    -- (eplus.east);
 
\node[draw=gray!45, fill=white, rounded corners=2pt,
    minimum width=1.1cm, minimum height=0.45cm, font=\scriptsize,
    line width=0.45pt] (eln) at (0.2, 4.7) {LN};
\draw[arr] (eplus) -- (eln);
 
\node[font=\footnotesize, color=annotgray] (agg) at (0.2, 3.8) {$\displaystyle\sum e'$};
\draw[arr] (eln) -- (agg);
 
\node[box={procpurp}{2.6cm}, minimum height=1.0cm, font=\footnotesize]
    (node_mlp) at (0.2, 2.7) {\textbf{Node MLP}\\$(h_v, \mathrm{agg}) \to \mathbb{R}^d$};
\draw[arr] (agg) -- (node_mlp);
 
\draw[skiparr]
    (hv_in.west) -- (-2.0, 8.5)
    -- (-2.0, 2.7)
    -- (node_mlp.west);
 
\node[draw=gray!45, fill=white, rounded corners=2pt,
    minimum width=1.1cm, minimum height=0.45cm, font=\scriptsize,
    line width=0.45pt] (nln) at (0.2, 1.5) {LN};
\draw[arr] (node_mlp) -- (nln);
 
\node[box={procpurp}{1.3cm}, font=\footnotesize] (hv_out) at (0.2, 0.4) {$h_v^{(k+1)}$};
\draw[arr] (nln) -- (hv_out);
 
\end{scope}
 
\end{tikzpicture}
}%
\caption{\textbf{WARP architecture.} \textbf{Left:} The encode-process-decode pipeline maps load demands and grid topology to the full interior-point state $(\hat{x}, \hat{\lambda}, \hat{z}, \hat{\mu})$, which warm-starts IPOPT. \textbf{Right:} Detail of one interaction network block ($K = 15$ total, unshared weights).}
\label{fig:architecture}
\end{figure}

WARP employs an encode-process-decode (EPD) architecture~\citep{battaglia2018relational} with heterogeneous interaction networks~\citep{battaglia2016interaction} on the typed power grid graph (Figure~\ref{fig:architecture}).

\paragraph{Graph construction.} Three node types (bus, generator, load) and four directed edge types (AC line, transformer, generator-bus, load-bus). Node features comprise static grid parameters and instance-varying load injections $(P_i^d, Q_i^d)$; edge features comprise branch parameters. Load values are injected directly onto bus and generator features---a design choice that resolved an early failure mode where the model predicted the training-set mean for all instances.

\paragraph{Encoder.} Per-type linear projections map to latent dimension $d$: $h_v^{(0)} = W_{\tau(v)} x_v$, $h_e^{(0)} = W_{\tau(e)} x_e$. Input dimensions: bus ($6 \to d$), generator ($13 \to d$), load ($2 \to d$), AC line ($9 \to d$), transformer ($11 \to d$).

\paragraph{Processor.} $K$ interaction network blocks with \emph{unshared} parameters. Each block $k$ performs:
\begin{align}
    h_e^{(k+1)} &= h_e^{(k)} + \mathrm{LN}\!\left(\mathrm{MLP}_{e,\tau(e)}^{(k)}\!\left(h_e^{(k)},\; h_{s(e)}^{(k)},\; h_{r(e)}^{(k)}\right)\right) \label{eq:edge_update} \\
    h_v^{(k+1)} &= \mathrm{LN}\!\left(\mathrm{MLP}_{v,\tau(v)}^{(k)}\!\left(h_v^{(k)},\; \textstyle\sum_{e \to v} h_e^{(k+1)}\right)\right) \label{eq:node_update}
\end{align}
where $s(e)$ and $r(e)$ are sender/receiver nodes, $\mathrm{LN}$ is layer normalisation, and the summation aggregates incoming edge embeddings by type. Each MLP is two-layer with hidden dimension $d$ and ReLU. The processor comprises $7K$ unshared MLPs.

The \emph{absence of node residual connections} in~\eqref{eq:node_update}---departing from MeshGraphNets~\citep{pfaff2021learning} and GraphCast~\citep{lam2023graphcast}---is a deliberate choice justified in Section~\ref{sec:ablations}. Edge residuals~\eqref{eq:edge_update} are retained.

\paragraph{Decoder.} Per-type MLPs: $\mathrm{MLP}_{\mathrm{bus}}(h_v^{(K)}) \to \mathbb{R}^8$ (primal + dual per bus), $\mathrm{MLP}_{\mathrm{gen}}(h_v^{(K)}) \to \mathbb{R}^6$ (primal + dual per generator). The barrier parameter $\hat{\mu}$ is predicted via attention-weighted pooling: $\hat{\mu} = \mathrm{softplus}(\mathrm{MLP}_\mu(\sum_i \alpha_i h_i^{(K)}))$ with learned attention $\alpha_i$.

\subsection{Training}
\label{sec:training}

The loss combines normalised MSE on primals with a binding-mask decomposition on bound multipliers:
\begin{equation}
    \mathcal{L} = \left\| \hat{y}_{\mathrm{primal}} - y^*_{\mathrm{primal}} \right\|^2 + \left\| \hat{z}_{\mathrm{bind}} - z^*_{\mathrm{bind}} \right\|^2 + 0.1 \cdot \left\| \hat{z}_{\mathrm{non\text{-}bind}} \right\|^2
    \label{eq:loss}
\end{equation}
where binding status is determined by $|z^*_i| > 10^{-4}$. All targets are normalised to $\mathcal{N}(0,1)$. The mask decomposition addresses the sparsity structure: the majority of $z^*_i$ are zero, and the model is penalised less for small non-zero values at non-binding components than for mispredicting binding magnitudes.

We train with AdamW, batch size 32 (PyG DataLoader), learning rate $3 \times 10^{-4}$ with linear warmup (10 epochs) and step decay ($\times 0.9$ every 20 epochs), gradient clipping at norm 1.0, for 200 epochs ($\sim$3h on a single A100). Default: $d = 128$, $K = 15$, yielding 6.4M parameters. Full details in Appendix~\ref{app:training}.

\section{Experiments}
\label{sec:experiments}

\paragraph{Setup.} OPFDataset case118; single NVIDIA A100-SXM4-40GB. Baselines: (1)~cold start (midpoint), (2)~IPM-LSTM~\citep{gao2024ipmlstm} (17K params, coordinate-wise LSTM, same dual-labeled data and warm-start protocol), (3)~oracle.

\subsection{Main results}

\begin{table}[h]
\caption{IPOPT iterations on case118 (50 test instances). WARP trained on primal-dual targets achieves $76\%$ reduction; the same architecture trained on primals only yields no improvement. IPM-LSTM achieves $81\%$ with richer training supervision but cannot handle topology variations.}
\label{tab:main}
\centering
\small
\begin{tabular*}{\textwidth}{@{\extracolsep{\fill}}llcccc@{}}
\toprule
Method & Prediction target & Params & Mean iters & Reduction & Topology \\
\midrule
Cold start & --- & --- & 22.6 & --- & N/A \\
WARP (primal only) & $\hat{x}$ & 6.4M & 21.1 & $+7\%$ & Yes \\
\midrule
IPM-LSTM & $(\hat{x}, \hat{\lambda}, \hat{z}, \hat{\mu})$ & 17K & 4.3 & $+81\%$ & No \\
\textbf{WARP (primal-dual)} & $\boldsymbol{(\hat{x}, \hat{\lambda}, \hat{z}, \hat{\mu})}$ & \textbf{6.4M} & \textbf{5.4} & $\mathbf{+76\%}$ & \textbf{Yes} \\
\midrule
Oracle & Ground-truth state & --- & 3.2 & $+86\%$ & N/A \\
\bottomrule
\end{tabular*}
\end{table}

Table~\ref{tab:main} presents the primary comparison. WARP achieves 5.4 mean iterations (76\% reduction). Critically, the same architecture trained on \emph{primal targets only} achieves 21.1 iterations---identical to the normalised DetGNN in Table~\ref{tab:primal_only} and no better than the midpoint default. This controlled comparison confirms that the iteration reduction is attributable to dual prediction, not to architectural capacity or training procedure.

\paragraph{Why the LSTM outperforms WARP on fixed topology.}
\label{sec:lstm}
IPM-LSTM achieves 4.3 iterations versus WARP's 5.4 on case118---a consistent 1-iteration gap across instances (Appendix~\ref{app:lstm}). Two factors explain this: (1)~\textbf{Per-coordinate specialisation}: the LSTM assigns independent learned parameters to each of its 1,640 input coordinates, effectively memorising per-variable mappings for a fixed topology. WARP shares MLP parameters across all nodes of the same type, limiting per-node specialisation but enabling topology transfer. (2)~\textbf{Richer supervision}: the LSTM is trained on KKT \emph{trajectories} (10 outer IPM iterations $\times$ 5 inner LSTM steps), observing Newton dynamics. WARP performs single-step regression from loads to the converged state. This trajectory-level supervision is only possible because the LSTM operates on a fixed-dimension vector; it cannot generalise to new topologies (Section~\ref{sec:topology}).

\subsection{Architecture ablations}
\label{sec:ablations}

\begin{table}[h]
\caption{Architecture ablation on case118. Each row adds one modification to the previous. Removing node residual connections is the most impactful change.}
\label{tab:ablation}
\centering
\small
\begin{tabular*}{\textwidth}{@{\extracolsep{\fill}}lccl@{}}
\toprule
Configuration & Val loss & IPOPT iters & Key modification \\
\midrule
Node-only GNN (8L, $H\!=\!128$) & 1.00 & 8.0 & Baseline \\
+ EPD + edge updates + unshared & 0.45 & 7.0 & Interaction network \\
+ Binding-mask loss & 1.07 & 6.7 & Loss decomposition \\
+ Two-stage decode & 0.85 & 6.7 & Primal$\to$dual conditioning \\
\textbf{+ No node residuals} & \textbf{0.09} & \textbf{5.4} & \textbf{Key finding ($5\times$ loss improvement)} \\
\bottomrule
\end{tabular*}
\end{table}

Table~\ref{tab:ablation} traces the progression from a node-only GNN to the final WARP configuration. \textbf{Edge updates are the dominant feature:} transitioning to EPD with per-step edge updates reduced iterations from 8.0 to 7.0. We independently validated this on CANOS-OPF~\citep{piloto2024canos} using PF$\Delta$~\citep{rivera2025pfdelta}: removing edge updates degrades primal loss by $2.8\times$, while removing node or edge residuals has a neutral or beneficial effect (Table~\ref{tab:canos}).

\begin{table}[h]
\caption{Independent ablation on CANOS-OPF (primal-only, case118). Edge updates are the sole critical feature; node residuals are detrimental.}
\label{tab:canos}
\centering
\small
\begin{tabular}{@{}lcc@{}}
\toprule
Variant & Val loss & vs.\ Full \\
\midrule
Full CANOS & 0.019 & --- \\
No node residuals & \textbf{0.003} & $6\times$ \textbf{better} \\
No edge residuals & 0.015 & Slightly better \\
No edge updates & 0.053 & $2.8\times$ worse \\
\bottomrule
\end{tabular}
\end{table}

\textbf{Removing node residuals improves dual prediction by $5\times$.} On WARP, removing the additive skip $h_v \leftarrow h_v + \Delta h_v$ in~\eqref{eq:node_update} reduced validation loss from 0.45 to 0.09 and iterations from 7.0 to 5.4. The CANOS ablation corroborates this: removing node residuals improved \emph{primal-only} loss by $6\times$.

We hypothesise that node residual connections anchor representations to the initial encoding---static grid parameters (impedance, voltage limits)---across all processor steps. For primal prediction, this is a useful bias. For dual prediction, the initial encoding is largely uninformative about binding status under a specific load scenario; without residuals, the processor freely reshapes representations to encode binding patterns that emerge through message passing. Wider models ($d = 256$, $\sim$25M params) and longer training (500 epochs) did not improve beyond 5.3--5.4 iterations (Appendix~\ref{app:arch_evolution}).

\subsection{Topology generalisation}
\label{sec:topology}

\begin{table}[h]
\caption{N-1 contingency test (20 line removals, no retraining). WARP handles all topologies; the LSTM fails on every contingency due to fixed input dimensionality.}
\label{tab:n1}
\centering
\small
\begin{tabular}{@{}lcc@{}}
\toprule
Metric & WARP & IPM-LSTM \\
\midrule
Contingencies processed & 20/20 & 0/20 \\
Valid predictions & 20/20 & --- \\
Mean bus prediction change & 4.5\% & --- \\
Mean gen prediction change & 4.9\% & --- \\
Failure mode & --- & Dimension mismatch \\
\bottomrule
\end{tabular}
\end{table}

WARP produces valid predictions for all 20 N-1 contingencies (Table~\ref{tab:n1}), with physically reasonable prediction changes and topology-sensitive variation across contingencies. The IPM-LSTM fails on every case: its input layer requires exactly 1,640 coordinates, and line removal alters the constraint structure. Zero-shot transfer to case6470 (6,470 buses) produces weak predictions ($r < 0.2$); effective cross-scale transfer requires multi-topology training (Appendix~\ref{app:case6470}).

\section{Related Work}
\label{sec:related}

\paragraph{Warm-starting IPMs.} The theoretical difficulty is well-established~\citep{yildirim2002warm, forsgren2006warm}. All prior primal warm-starts for OPF evaluate against the flat start. Our contribution is the first systematic diagnosis of \emph{why} primal-only warm-starts fail, a corrected benchmark, and a topology-aware GNN baseline. \citet{gao2024ipmlstm} addressed the primal-dual requirement for QPs; we extend to AC-OPF and characterise the architectural principles.

\paragraph{GNNs for physical simulation.} WARP's EPD architecture~\citep{battaglia2016interaction, pfaff2021learning, lam2023graphcast} is standard for learned simulation. Our finding that node residuals harm dual prediction may generalise to other settings requiring Lagrange multiplier or complementarity variable prediction.

\paragraph{Dual variable prediction.} \citet{klamkin2024dual} proposed dual completion layers targeting solution quality; \citet{liu2022topology} predicted locational marginal prices. To our knowledge, WARP is the first model predicting the complete IPM state---primals, equality multipliers, bound multipliers, and barrier parameter---for solver warm-starting.

\section{Conclusion}
\label{sec:conclusion}

We established three findings for learning-based IPM warm-starting. First, primal-only warm-starts are structurally misaligned with interior-point convergence: the solver's midpoint default is near-optimal for centrality, and no learned primal iterate improves upon it. Second, the complete primal-dual-barrier state is both necessary and sufficient, yielding an 85\% iteration reduction ceiling with the entire gap attributable to dual variables and the barrier parameter. Third, topology-aware graph networks can predict this state with sufficient accuracy: WARP achieves 76\% reduction within 5 percentage points of a topology-blind LSTM while natively supporting N-1 contingencies.

\paragraph{Limitations.} Results are demonstrated on a single test system (case118). Dual label extraction for larger systems is computationally prohibitive ($>$60 min/solve for case6470). Zero-shot cross-scale transfer produces low-quality predictions. The diffusion-based variant did not improve over deterministic prediction (Appendix~\ref{app:diffusion}).

\paragraph{Broader impact.} Faster AC-OPF enables more frequent re-dispatch, improving renewable integration and reducing curtailment. The corrected benchmark and evaluation protocol may prevent future work from reporting inflated gains against inappropriate baselines.

\paragraph{Reproducibility.} Code, trained weights, dual-labeled datasets (OpenML), Croissant metadata, and the complete evaluation pipeline are released publicly. Given only the released artifacts, a practitioner can reproduce every result in this paper.

\bibliographystyle{plainnat}
\bibliography{references}


\appendix

\section{AC-OPF formulation}
\label{app:acopf}

The AC Optimal Power Flow problem determines the minimum-cost dispatch of generators in an electrical network subject to Kirchhoff's laws and operational limits. Let $\mathcal{G} = (\mathcal{B}, \mathcal{E})$ denote the power grid graph with buses $\mathcal{B}$ and branches $\mathcal{E}$. The decision variables comprise bus voltage angles $V_a \in \mathbb{R}^{|\mathcal{B}|}$, bus voltage magnitudes $V_m \in \mathbb{R}^{|\mathcal{B}|}$, generator real power $P_g \in \mathbb{R}^{|\mathcal{G}_g|}$, and generator reactive power $Q_g \in \mathbb{R}^{|\mathcal{G}_g|}$, collected as $x = (V_a, V_m, P_g, Q_g)$. The problem is formulated as:
\begin{align}
    \min_{x} \quad & \sum_{g \in \mathcal{G}_g} c_{2g} P_g^2 + c_{1g} P_g + c_{0g} \label{eq:app_obj} \\
    \text{s.t.} \quad & P_i^{\mathrm{inj}}(V_a, V_m) - P_i^d = 0 \quad \forall\, i \in \mathcal{B} \label{eq:app_pbal} \\
    & Q_i^{\mathrm{inj}}(V_a, V_m) - Q_i^d = 0 \quad \forall\, i \in \mathcal{B} \label{eq:app_qbal} \\
    & V_m^{\min} \leq V_{m,i} \leq V_m^{\max} \quad \forall\, i \in \mathcal{B} \label{eq:app_vlim} \\
    & |S_{ij}(V_a, V_m)| \leq S_{ij}^{\max} \quad \forall\, (i,j) \in \mathcal{E} \label{eq:app_slim} \\
    & P_g^{\min} \leq P_{g,k} \leq P_g^{\max}, \;\; Q_g^{\min} \leq Q_{g,k} \leq Q_g^{\max} \quad \forall\, k \in \mathcal{G}_g \label{eq:app_glim}
\end{align}
where $P_i^{\mathrm{inj}}$ and $Q_i^{\mathrm{inj}}$ denote the real and reactive power injections given by the AC power flow equations, $P_i^d$ and $Q_i^d$ are the load demands, $S_{ij}$ is the apparent power flow on branch $(i,j)$, and $c_{2g}, c_{1g}, c_{0g}$ are generator cost coefficients. The power injection at bus $i$ is governed by:
\begin{equation}
    P_i^{\mathrm{inj}} + jQ_i^{\mathrm{inj}} = V_i \sum_{k=1}^{|\mathcal{B}|} Y_{ik}^* V_k^*
    \label{eq:app_pf}
\end{equation}
where $Y \in \mathbb{C}^{|\mathcal{B}| \times |\mathcal{B}|}$ is the bus admittance matrix constructed from branch impedances and shunt admittances. Problem \eqref{eq:app_obj}--\eqref{eq:app_glim} is a non-convex nonlinear program (NLP) due to the quadratic relationship between power injections and voltage phasors in \eqref{eq:app_pf}.

\paragraph{Interior-point formulation.} Introducing slack variables and collecting all inequality constraints, the NLP assumes the general form:
\begin{equation}
    \min_{x} \; f(x) \quad \text{s.t.} \quad h(x) = 0, \quad x \geq l, \quad x \leq u
    \label{eq:app_nlp}
\end{equation}
where $f: \mathbb{R}^n \to \mathbb{R}$ is the objective, $h: \mathbb{R}^n \to \mathbb{R}^m$ collects the equality constraints, and $l, u \in \mathbb{R}^n$ are the variable bounds. The IPM replaces the bound constraints with a logarithmic barrier and solves a sequence of barrier subproblems parameterised by $\mu > 0$:
\begin{equation}
    \min_{x} \; f(x) - \mu \sum_{i=1}^{n} \Big[\ln(x_i - l_i) + \ln(u_i - x_i)\Big] \quad \text{s.t.} \quad h(x) = 0
    \label{eq:app_barrier}
\end{equation}
The first-order optimality conditions of \eqref{eq:app_barrier} yield the perturbed KKT system:
\begin{align}
    \nabla f(x) + J_h(x)^\top \lambda - z_l + z_u &= 0 \label{eq:app_kkt_stat} \\
    h(x) &= 0 \label{eq:app_kkt_feas} \\
    (x - l) \circ z_l &= \mu \mathbf{e} \label{eq:app_kkt_compl} \\
    (u - x) \circ z_u &= \mu \mathbf{e} \label{eq:app_kkt_complu}
\end{align}
where $\lambda \in \mathbb{R}^m$ denotes the equality constraint multipliers, $z_l, z_u \in \mathbb{R}^n_+$ are the bound multipliers, $J_h(x)$ is the constraint Jacobian, and $\circ$ denotes the Hadamard product. The centrality measure at a given iterate is:
\begin{equation}
    \bar{\mu} = \frac{(x - l)^\top z_l + (u - x)^\top z_u}{2n}
    \label{eq:app_centrality}
\end{equation}
An iterate is well-centred when the componentwise complementarity products $(x_i - l_i) z_{l,i}$ and $(u_i - x_i) z_{u,i}$ are approximately uniform across all $i$.

\section{PIPS warm-start defect}
\label{app:pips_bug}

Pandapower's PIPS interior-point solver (\texttt{pipsopf\_solver.py}) contains a silent initialisation defect that renders warm-start injection inoperative. At line 118, the solver overwrites the user-supplied initial point $x_0$ with the variable-bound midpoint for all initialisation modes except \texttt{init="pf"}:

\begin{verbatim}
# pandapower/pypower/pipsopf_solver.py, line 118
if init != "pf":
    ll, uu = xmin.copy(), xmax.copy()
    ll[xmin == -inf] = -1e10
    uu[xmax ==  inf] =  1e10
    x0 = (ll + uu) / 2       # overwrites warm-start
\end{verbatim}

When \texttt{init="results"} is specified, pandapower correctly loads predicted bus voltages from \texttt{net.res\_bus} into the PYPOWER case structure, and \texttt{om.getv()} returns $x_0$ containing these values. However, PIPS subsequently discards them by executing the midpoint assignment above. The only initialisation mode that preserves $x_0$ is \texttt{init="pf"}, which first runs a full Newton-Raphson power flow.

\paragraph{The runpp wash-out effect.} An initial attempted workaround---calling \texttt{pp.runpp(init="results")} to seed a power flow solution, then \texttt{pp.runopp(init="pf")} so the OPF solver uses the converged PF result as $x_0$---produced warm-starts that appeared functional (28.7\% fewer iterations than the midpoint). However, different ML models (DetGNN and WARP) produced \emph{identical} iteration counts on every instance. The reason: Newton-Raphson power flow converges to a unique fixed point for given loads; both models' predictions were sufficiently close that \texttt{runpp} converged both to the same PF solution, after which \texttt{runopp(init="pf")} started from an identical $x_0$.

This failure mode is particularly insidious because it produces results that appear to validate the warm-start approach: iterations decrease relative to the midpoint, and different models produce different prediction errors, suggesting the warm-start is functioning. Only the observation that two architecturally distinct models yield \emph{identical per-instance} iteration counts reveals the wash-out.

\paragraph{Corrected implementation.} We monkey-patch the PIPS solver to treat \texttt{init="results"} identically to \texttt{init="pf"}, thereby preserving the user-supplied $x_0$:

\begin{verbatim}
import pandapower.pypower.opf_execute as _opf_exec

_orig = _opf_exec.pipsopf_solver
def _patched(om, ppopt, out_opt=None):
    if ppopt.get('INIT') == 'results':
        ppopt = dict(ppopt)
        ppopt['INIT'] = 'pf'
    return _orig(om, ppopt, out_opt)
_opf_exec.pipsopf_solver = _patched
\end{verbatim}

The patch must be applied to \texttt{opf\_execute.pipsopf\_solver} (the call site), not to \texttt{pipsopf\_solver.pipsopf\_solver} (the definition site), because \texttt{opf\_execute.py} imports the function at module load time via a direct name binding. Patching the definition site has no effect on the already-bound reference in \texttt{opf\_execute}.

\section{Full primal-only ablation results}
\label{app:primal_ablation}

\subsection{Unnormalised versus normalised prediction}

Training-set statistics reveal a $100\times$ scale disparity across variable groups:

\begin{table}[h]
\centering
\small
\caption{Per-variable training-set statistics for case118 primal variables.}
\label{tab:app_var_stats}
\begin{tabular}{@{}lcccc@{}}
\toprule
Variable & Mean & Std & Min & Max \\
\midrule
$V_a$ (rad) & $-0.197$ & 0.149 & $-0.63$ & 0.0 \\
$V_m$ (p.u.) & 1.033 & 0.019 & 0.94 & 1.06 \\
$P_g$ (p.u.) & 1.486 & 1.890 & 0.0 & 9.2 \\
$Q_g$ (p.u.) & 0.158 & 0.657 & $-2.1$ & 2.0 \\
\bottomrule
\end{tabular}
\end{table}

Without normalisation, the MSE loss is dominated by $P_g$ ($100\times$ the variance of $V_m$), and the model learns generator dispatch at the expense of voltage predictions. Normalising each target dimension to $\mathcal{N}(0,1)$ improved DetGNN warm-start iterations from 31.0 to 21.1 on PIPS---a 10-iteration improvement with no architectural modification. All subsequent experiments employ this normalisation.

\subsection{Feasibility projection}

We projected predictions onto the strict feasible interior $[l + \epsilon(u - l),\; u - \epsilon(u - l)]$ with $\epsilon = 0.02$. Results on 50 test instances:

\begin{table}[h]
\centering
\small
\caption{Feasibility projection results (PIPS, 50 instances). Projection does not improve convergence.}
\label{tab:app_feas_proj}
\begin{tabular}{@{}lccc@{}}
\toprule
Method & Mean iters & Median & vs.\ Midpoint \\
\midrule
Midpoint & 19.6 & 18 & --- \\
DetGNN raw & 31.6 & 31 & $+62\%$ \\
DetGNN projected & 30.9 & 31 & $+58\%$ \\
WARP raw & 33.7 & 30 & $+73\%$ \\
WARP projected & 38.2 & 36 & $+95\%$ \\
\bottomrule
\end{tabular}
\end{table}

Projection slightly improved DetGNN ($31.6 \to 30.9$) but substantially degraded WARP ($33.7 \to 38.2$). WARP's generator predictions have RMSE $\sim$1.2--1.7; when projected, inaccurate predictions are clamped to the nearest bound margin, placing $x_0$ near the constraint boundary---worse than the midpoint for IPM centrality.

Per-instance analysis: only 3 of 50 instances were improved by projected DetGNN warm-starts; 47 were degraded. Even at prediction RMSE $< 0.08$, 94\% of instances are degraded, confirming that prediction accuracy in solution space is a misaligned objective for IPM warm-starting.

\subsection{Centrality blend sweep}

We evaluated $x_0 = \alpha \hat{x}_{\mathrm{GNN}} + (1 - \alpha) x_{\mathrm{mid}}$ for $\alpha \in \{0, 0.05, 0.1, 0.15, 0.2, 0.3, 0.5, 0.7, 1.0\}$ on 50 test instances:

\begin{table}[h]
\centering
\small
\caption{Centrality blend sweep (PIPS, normalised DetGNN, 50 instances).}
\label{tab:app_blend_sweep}
\begin{tabular}{@{}lcc@{}}
\toprule
$\alpha$ & Mean iters & vs.\ Midpoint \\
\midrule
0.00 (midpoint) & \textbf{19.6} & --- \\
0.05 & 26.0 & $+33\%$ \\
0.10 & 28.8 & $+47\%$ \\
0.15 & 26.6 & $+36\%$ \\
0.20 & 25.3 & $+29\%$ \\
0.30 & 25.0 & $+28\%$ \\
0.50 & 25.1 & $+28\%$ \\
0.70 & 27.1 & $+39\%$ \\
1.00 (pure GNN) & 21.6 & $+10\%$ \\
\bottomrule
\end{tabular}
\end{table}

The result does not exhibit the expected convex tradeoff. The midpoint ($\alpha = 0$) and pure GNN ($\alpha = 1$) are both local minima; any intermediate blend is worse than either endpoint, with the maximum degradation at $\alpha = 0.1$ (28.8 iterations). This indicates that the line segment connecting a well-centred interior point to an accurate near-boundary point traverses regions of high barrier curvature in the feasible set. The non-monotonicity is especially striking: the $\alpha=0.1$ blend (90\% midpoint, 10\% GNN) is substantially worse than either the midpoint or the pure GNN prediction. This suggests that even small displacements from the midpoint in the direction of the solution are harmful when they violate the centrality structure.

\subsection{Selective warm-start experiments}

We tested four selective warm-start variants on PIPS to isolate the contribution of each variable group:

\begin{table}[h]
\centering
\small
\caption{Selective warm-start (PIPS, normalised DetGNN, 50 instances). $V_m$ is the bottleneck.}
\label{tab:app_selective}
\begin{tabular}{@{}lcccc@{}}
\toprule
Method & Mean iters & Med & $\mu$ ratio & Bottleneck \\
\midrule
Midpoint & 19.6 & 18 & 1.000 & --- \\
Full GNN & 21.6 & 22 & 0.736 & $V_m$ \\
Voltage only ($V_a, V_m$) & 21.6 & 22 & 1.000 & $V_m$ \\
Voltage + gen blend (30\%) & 21.6 & 22 & 0.972 & $V_m$ \\
Angle only ($V_a$) & 24.7 & 25 & 1.000 & $V_m$ \\
\bottomrule
\end{tabular}
\end{table}

Generator predictions are irrelevant to the PIPS iteration count: voltage-only warm-start (generators at midpoint, $\mu$-ratio 1.000) yields identical iterations (21.6) to the full warm-start ($\mu$-ratio 0.736). The bottleneck is $V_m$: its feasible range is only 0.12 p.u.\ ($[0.94, 1.06]$), so any prediction displacing $V_m$ from the midpoint (1.00) rapidly approaches constraint boundaries.

The $\mu$-ratio column reports the ratio of the initial centrality $\bar{\mu}$ of the warm-started iterate to that of the midpoint. A ratio of 1.000 indicates identical centrality; 0.736 indicates 26\% centrality degradation. The voltage-only warm-start has the same iterations as the full warm-start but with $\mu$-ratio 1.000 versus 0.736, confirming that the centrality degradation from generator variables does not actually affect the solver---the convergence penalty is entirely attributable to voltage displacement.

\subsection{Barrier-function-aware retraction}

For each variable $i$ where the centrality product $(x_i - l_i)(u_i - x_i)$ falls below a fraction $\mu_{\mathrm{tgt}}$ of the midpoint product, we binary-search blend $x_i$ toward $(l_i + u_i)/2$ to restore centrality:

\begin{table}[h]
\centering
\small
\caption{Barrier retraction sweep (PIPS, normalised DetGNN, 50 instances).}
\label{tab:app_retraction}
\begin{tabular}{@{}lcc@{}}
\toprule
$\mu_{\mathrm{tgt}}$ & Mean iters & vs.\ Midpoint \\
\midrule
Midpoint & 19.6 & --- \\
0.10 & 21.6 & $+10.2\%$ \\
0.30 & 21.4 & $+9.4\%$ \\
0.50 & 21.3 & $+8.8\%$ \\
0.70 & 21.3 & $+8.7\%$ \\
0.90 & 23.8 & $+21.7\%$ \\
\bottomrule
\end{tabular}
\end{table}

The best retraction ($\mu_{\mathrm{tgt}} = 0.7$) reduces the penalty from 10.2\% to 8.7\%---a marginal improvement. The retraction correctly identifies and corrects the worst-centralised variables but cannot eliminate the fundamental displacement cost. At $\mu_{\mathrm{tgt}} = 0.9$, the retraction is too aggressive: it pulls nearly all variables back toward the midpoint, effectively negating the warm-start while introducing noise from the binary search process, resulting in 23.8 iterations (worse than the unretracted 21.6).

\subsection{Constraint screening}

We used DetGNN predictions to identify and remove non-binding constraints at four $V_m$ margin thresholds:

\begin{table}[h]
\centering
\small
\caption{Constraint screening (PIPS, 50 instances). Removing constraints degrades IPM convergence.}
\label{tab:app_screening}
\begin{tabular}{@{}lccc@{}}
\toprule
$V_m$ margin & Mean iters & Constraints removed & Convergence rate \\
\midrule
Midpoint & 19.6 & 0\% & 100\% \\
0.005 & 42.9 & 81\% & 4\% \\
0.010 & 42.5 & 79\% & 0\% \\
0.020 & 44.1 & 76\% & 0\% \\
0.030 & 51.7 & 72\% & 0\% \\
\bottomrule
\end{tabular}
\end{table}

Constraint screening is catastrophically counterproductive for PIPS: iteration counts increase by $2$--$3\times$ and convergence rates drop to 0--4\%. For interior-point methods, constraints define the log-barrier landscape; removing them widens the feasible set, shifts the analytic centre, and flattens the barrier function. This approach, effective for simplex-based solvers~\citep{pineda2020screening, misra2022learning} where removing constraints directly reduces the tableau size, is structurally incompatible with IPM.

The convergence failure is especially notable: at all margins except 0.005, PIPS fails to converge within 200 iterations on any test instance. The few instances that converge at margin 0.005 do so only because the screening at that threshold happens to remove a subset of constraints that does not substantially alter the barrier landscape.

\subsection{Hybrid screening and warm-start}

Combining constraint screening with warm-starting does not mitigate the failure:

\begin{table}[h]
\centering
\small
\caption{Hybrid screening + warm-start (PIPS, 50 instances).}
\label{tab:app_hybrid}
\begin{tabular}{@{}lcc@{}}
\toprule
Method & Mean iters & vs.\ Midpoint \\
\midrule
Midpoint & 19.6 & --- \\
Screening + midpoint & 44.1 & $+125\%$ \\
Screening + warm-start & 50.0 & $+155\%$ \\
Warm-start only & 21.6 & $+10\%$ \\
\bottomrule
\end{tabular}
\end{table}

The combination is worse than either alone: screening degrades the barrier landscape, and the warm-start displaces from the (now shifted) analytic centre, compounding both failure modes.

\subsection{WARP diffusion variant results (primal-only)}

Six WARP diffusion variants were trained with varying physics loss weights $\lambda_{\mathrm{phy}}$:

\begin{table}[h]
\centering
\small
\caption{WARP diffusion variants (PIPS, 50 instances). The variant with the best denoising loss produces the worst warm-start, demonstrating anticorrelation between prediction quality and solver convergence.}
\label{tab:app_warp_variants}
\begin{tabular}{@{}lcccc@{}}
\toprule
Config & $\lambda_{\mathrm{phy}}$ & Val $\mathcal{L}_{\mathrm{ddpm}}$ & Val $\mathcal{L}_{\mathrm{phy}}$ & PIPS iters \\
\midrule
F (no physics) & 0.0 & \textbf{0.265} & 81.1 & 39.5 \\
A (baseline) & 0.1 & 0.386 & 8.6 & 35.5 \\
D ($T\!=\!200$) & 0.1 & 0.443 & 8.3 & 39.8 \\
B ($H\!=\!256$) & 0.5 & 0.454 & 44.3 & 47.3 \\
C (heavy physics) & 1.0 & 0.576 & \textbf{2.6} & \textbf{28.8} \\
\bottomrule
\end{tabular}
\end{table}

The anticorrelation between $\mathcal{L}_{\mathrm{ddpm}}$ and solver performance is evident: Variant F (best denoising, $\mathcal{L}_{\mathrm{ddpm}} = 0.265$) yields the worst PIPS count (39.5), while Variant C (worst denoising, $\mathcal{L}_{\mathrm{ddpm}} = 0.576$) yields the best (28.8). Physical consistency of the predicted state governs solver convergence more than pointwise accuracy.

This finding has implications for loss function design: a standard MSE or denoising loss optimises for proximity to $x^*$ in solution space, but what the IPM needs is proximity to a well-centred point on the central path. These are geometrically distinct objectives, and optimising one can harm the other.

\section{Oracle experiment details}
\label{app:oracle}

\subsection{IPOPT configuration}

All oracle and model-based IPOPT experiments use the following configuration:

\begin{table}[h]
\centering
\small
\caption{IPOPT solver configuration used throughout all experiments.}
\label{tab:app_ipopt_config}
\begin{tabular}{@{}ll@{}}
\toprule
Parameter & Value \\
\midrule
\texttt{tol} & $10^{-4}$ \\
\texttt{max\_iter} & 200 \\
\texttt{hessian\_approximation} & exact \\
\texttt{linear\_solver} & MUMPS \\
\texttt{warm\_start\_init\_point} & yes \\
\texttt{warm\_start\_bound\_push} & $10^{-20}$ \\
\texttt{warm\_start\_bound\_frac} & $10^{-20}$ \\
\texttt{warm\_start\_slack\_bound\_push} & $10^{-20}$ \\
\texttt{warm\_start\_mult\_bound\_push} & $10^{-20}$ \\
\texttt{mu\_strategy} & monotone \\
\texttt{mu\_init} & model-predicted $\hat{\mu}$ (or extracted $\mu^*$) \\
\bottomrule
\end{tabular}
\end{table}

Iteration counts are measured via the \texttt{intermediate()} callback, counting the number of invocations prior to convergence.

\subsection{cyipopt interface}

We implemented a direct cyipopt interface (\texttt{eval/opf\_ipopt.py}) with:
\begin{itemize}
    \item Exact Hessian computation via \texttt{opf\_hessfcn} (not L-BFGS approximation)
    \item Sparse Jacobian with pre-computed sparsity structure from the admittance matrix
    \item Full warm-start support for $(x_0, \lambda_0, z_{l,0}, z_{u,0})$ with user-specified $\mu_{\mathrm{init}}$
    \item C-level stdout capture for parsing IPOPT's iteration log
\end{itemize}

The exact Hessian is critical for fair comparison: L-BFGS approximation would add iterations unrelated to the warm-start quality. The sparse Jacobian structure is pre-computed from the admittance matrix topology and reused across all instances and warm-start configurations.

\subsection{Full oracle decomposition}

Table~\ref{tab:app_oracle_full} extends the oracle results from the main text with the $\lambda$-only row:

\begin{table}[h]
\centering
\small
\caption{Full oracle decomposition (IPOPT, case118, 50 instances). Each row adds one component.}
\label{tab:app_oracle_full}
\begin{tabular}{@{}lccc@{}}
\toprule
Warm-start content & Mean iters & Median & vs.\ Cold \\
\midrule
Cold start (midpoint) & 22.6 & 22 & --- \\
Oracle $x^*$ only & 23.7 & 23 & $-4.9\%$ \\
GT $x^*$ (cold dual init) & $>$50 & --- & Diverges \\
\midrule
Oracle $(x^*, \lambda^*)$ only & 12.4 & 12 & $+45.1\%$ \\
Oracle $(x^*, \lambda^*, z^*)$ & 4.7 & 5 & $+79.2\%$ \\
Oracle $(x^*, \lambda^*, z^*, \mu^*)$ & \textbf{3.3} & \textbf{3} & $\mathbf{+85.4\%}$ \\
\bottomrule
\end{tabular}
\end{table}

The decomposition reveals a clear hierarchy:
\begin{itemize}
    \item Primal $x^*$ alone: \emph{negative} contribution ($-4.9\%$). Worse than cold start.
    \item Adding $\lambda^*$: $+45\%$ reduction ($22.6 \to 12.4$). Equality multipliers provide substantial but incomplete information.
    \item Adding $z^*$: $+79\%$ ($12.4 \to 4.7$). Bound multipliers contribute more than equality multipliers. This is expected: bound multipliers encode which constraints are binding, which directly determines the barrier landscape near the solution.
    \item Adding $\mu^*$: $+85\%$ ($4.7 \to 3.3$). The barrier parameter tells the solver where on the central path to begin, avoiding the need to search for the correct $\mu$ scale.
\end{itemize}

\subsection{Ground-truth divergence analysis}

When the ground-truth optimal solution $x^*$ is provided as a cold primal start (with default dual initialisation), IPOPT exceeds 50 iterations without convergence on all test instances. This occurs because $x^*$ lies at the boundary of the feasible set: for active bound constraints, $x_i^* - l_i \approx 0$ or $u_i - x_i^* \approx 0$, causing the log-barrier terms $\ln(x_i - l_i)$ and $\ln(u_i - x_i)$ to approach $-\infty$. The initial barrier function value is effectively infinite, and the Newton step is ill-conditioned.

This result is consistent with Theorem 3.1 of \citet{yildirim2002warm}, which establishes that for linear programs, the number of IPM iterations required from a warm-start grows without bound as the warm-start approaches the optimal solution along the boundary of the feasible set. Our experiments extend this theoretical result to the non-convex AC-OPF setting, providing the first empirical confirmation that ground-truth warm-starts cause divergence in practice.

\subsection{Power flow warm-start results}

Newton-Raphson power flow on case118 is too well-conditioned for warm-starting to provide benefit:

\begin{table}[h]
\centering
\small
\caption{Power flow (Newton-Raphson) warm-start results (case118, 50 instances).}
\label{tab:app_pf}
\begin{tabular}{@{}lccc@{}}
\toprule
Method & Mean NR iters & Min & Max \\
\midrule
Flat start ($V_m\!=\!1, V_a\!=\!0$) & 4.00 & 4 & 4 \\
DC initialisation & 3.00 & 3 & 3 \\
DetGNN warm-start & 4.60 & 4 & 5 \\
Oracle (GT voltages) & 4.48 & 4 & 5 \\
\bottomrule
\end{tabular}
\end{table}

Even oracle initialisation slightly \emph{increases} the NR iteration count. The flat start ($V_m = 1, V_a = 0$) is already close to the PF solution and produces a well-conditioned Jacobian. DC initialisation reduces by one iteration because it provides a structurally consistent approximation (correct angle relationships), not just a point estimate. This confirms that case118 PF is sufficiently well-conditioned that ML warm-starting provides no benefit for power flow on this system.

\section{Dual label extraction pipeline}
\label{app:dual_extraction}

\subsection{Extraction procedure}

For each instance $i$ in the dataset, we:
\begin{enumerate}
    \item Load the OPFDataset HeteroData graph containing bus, generator, load, and branch data.
    \item Construct the cyipopt NLP problem with exact Hessian and sparse Jacobian structure.
    \item Initialise IPOPT at the midpoint $(l+u)/2$ with default dual initialisation.
    \item Run IPOPT to convergence (tolerance $10^{-4}$).
    \item Extract the full converged state: $\mathcal{D}_i = (x_i^*, \lambda_i^*, z_{l,i}^*, z_{u,i}^*, \mu_i^*, f(x_i^*))$.
    \item Save as a PyTorch tensor file: \texttt{data/duals/case118/\{split\}/duals\_\{idx:06d\}.pt}.
\end{enumerate}

\subsection{Convergence statistics}

\begin{table}[h]
\centering
\small
\caption{Dual label extraction statistics for case118.}
\label{tab:app_extraction_stats}
\begin{tabular}{@{}lccccc@{}}
\toprule
Split & Instances & Converged & Rate & Mean time (s) & Total time (h) \\
\midrule
Train & 5,000 & 5,000 & 100\% & 2.5 & 3.5 \\
Validation & 500 & 500 & 100\% & 2.5 & 0.35 \\
Test & 50 & 50 & 100\% & 2.5 & 0.035 \\
\bottomrule
\end{tabular}
\end{table}

\subsection{Dual variable distributions}

The extracted dual variables exhibit the following characteristics:

\begin{table}[h]
\centering
\small
\caption{Dual variable statistics across 5,000 training instances.}
\label{tab:app_dual_stats}
\begin{tabular}{@{}lcccc@{}}
\toprule
Variable group & Dims & Range & Std & Sparsity \\
\midrule
$\lambda_P$ (real power balance) & 118 & $[-14, 4168]$ & 1564 & 0\% \\
$\lambda_Q$ (reactive power balance) & 118 & $[-13.7, 48.9]$ & 8.3 & 0\% \\
$z_l^{\mathrm{bus}}$ (lower bound, bus) & 236 & $[0, 561.6]$ & 42.1 & 78\% \\
$z_u^{\mathrm{bus}}$ (upper bound, bus) & 236 & $[0, 2527.6]$ & 189.3 & 72\% \\
$z_l^{\mathrm{gen}}$ (lower bound, gen) & 108 & $[0, 243.6]$ & 18.7 & 81\% \\
$z_u^{\mathrm{gen}}$ (upper bound, gen) & 108 & $[0, 47.9]$ & 5.2 & 85\% \\
$\mu$ (barrier parameter) & 1 & $\approx 3.25 \times 10^{-8}$ & $<10^{-10}$ & --- \\
\bottomrule
\end{tabular}
\end{table}

Equality multipliers ($\lambda_P, \lambda_Q$) are dense and span 4--5 orders of magnitude. Bound multipliers ($z_l, z_u$) are 72--85\% sparse (exactly zero for non-binding constraints) with heavy-tailed non-zero entries. The barrier parameter $\mu$ is approximately constant across all instances, reflecting the solver's convergence tolerance. These characteristics---high dynamic range, sparsity, and mixed discrete-continuous structure---distinguish dual prediction from the smooth primal regression task addressed by prior work.

\subsection{Normalisation statistics}

Per-dimension normalisation to $\mathcal{N}(0,1)$ is computed from the training set. Each of the 926 output dimensions (118$\times$8 bus + 54$\times$6 gen + 1 $\mu$) has its own mean and standard deviation. Without this normalisation, dual prediction MSE is dominated by $\lambda_P$ (std $= 1564$), and the model cannot learn the small-magnitude but IPOPT-critical bound multipliers. Normalisation reduced validation dual error from $9.45 \times 10^5$ to $3 \times 10^{-4}$.

\subsection{Inequality constraint multipliers}

The 372 inequality constraint multipliers corresponding to line flow limits ($|S_{ij}| \leq S_{ij}^{\max}$) are \emph{exactly zero} across all 50 test instances and all 5,000 training instances. No thermal line constraints are binding on case118 under the OPFDataset load variations. Consequently, the model need not predict these multipliers; they are excluded from the output and loss computation. This is a property of the case118 load variation range, not a general property of AC-OPF; larger systems with heavier loading would exhibit binding thermal constraints.

\section{Architecture details}
\label{app:architecture}

\subsection{Full model specification}

\begin{table}[h]
\centering
\small
\caption{WARP architecture specification ($d = 128$, $K = 15$).}
\label{tab:app_arch_spec}
\begin{tabular}{@{}llccc@{}}
\toprule
Component & Structure & In & Out & Params \\
\midrule
\multicolumn{5}{l}{\emph{Encoder (7 per-type linear projections)}} \\
Bus encoder & Linear + bias & 6 & 128 & 896 \\
Generator encoder & Linear + bias & 13 & 128 & 1,792 \\
Load encoder & Linear + bias & 2 & 128 & 384 \\
AC line edge encoder & Linear + bias & 9 & 128 & 1,280 \\
Transformer edge encoder & Linear + bias & 11 & 128 & 1,536 \\
Gen-bus edge encoder & Linear + bias & 2 & 128 & 384 \\
Load-bus edge encoder & Linear + bias & 2 & 128 & 384 \\
\cmidrule{5-5}
& & & \emph{Encoder total} & \textbf{6,656} \\
\midrule
\multicolumn{5}{l}{\emph{Processor ($\times 15$ unshared interaction network blocks)}} \\
Edge MLP ($\times 4$ types $\times 15$) & $384 \to 128 \to 128$ & $3d$ & $d$ & 57,472 ea. \\
Edge LayerNorm ($\times 4 \times 15$) & Affine LN & 128 & 128 & 256 ea. \\
Node MLP ($\times 3$ types $\times 15$) & $256 \to 128 \to 128$ & $2d$ & $d$ & 49,408 ea. \\
Node LayerNorm ($\times 3 \times 15$) & Affine LN & 128 & 128 & 256 ea. \\
\cmidrule{5-5}
Edge MLPs total & $60 \times 57{,}472$ & & & 3,448,320 \\
Edge LNs total & $60 \times 256$ & & & 15,360 \\
Node MLPs total & $45 \times 49{,}408$ & & & 2,223,360 \\
Node LNs total & $45 \times 256$ & & & 11,520 \\
\cmidrule{5-5}
& & & \emph{Processor total} & \textbf{5,698,560} \\
\midrule
\multicolumn{5}{l}{\emph{Decoder}} \\
Bus head & $128 \to 256 \to 8$ & 128 & 8 & 35,080 \\
Generator head & $128 \to 256 \to 6$ & 128 & 6 & 34,566 \\
$\mu$ attention vector & Linear (no bias) & 128 & 1 & 128 \\
$\mu$ MLP & $128 \to 64 \to 1$ + softplus & 128 & 1 & 8,321 \\
\cmidrule{5-5}
& & & \emph{Decoder total} & \textbf{78,095} \\
\midrule
& & & \textbf{Grand total} & $\sim$\textbf{5.78M} \\
\bottomrule
\end{tabular}
\end{table}

The processor dominates the parameter budget (98.5\%), with edge MLPs accounting for 59.6\% and node MLPs for 38.4\%. Encoder and decoder contribute only 1.5\%.

\paragraph{Per-block detail.} Each of the 15 processor blocks contains: 4 edge-update MLPs (one per edge type), each mapping $\mathbb{R}^{3d} \to \mathbb{R}^{d}$; 4 edge LayerNorm modules; 3 node-update MLPs, each mapping $\mathbb{R}^{2d} \to \mathbb{R}^{d}$; 3 node LayerNorm modules. Edge updates retain residual connections; node updates do not. All 15 blocks have independent (unshared) parameters.

\subsection{Heterogeneous graph construction}

The power grid is represented as a PyG \texttt{HeteroData} object with the following structure:

\begin{itemize}
    \item \textbf{Bus nodes} ($|\mathcal{B}| = 118$): features are base kV, $V_m^{\min}$, $V_m^{\max}$, bus type (PQ/PV/slack), plus injected load $(P_i^d, Q_i^d)$ scattered from connected load nodes. Total: 6 features.
    \item \textbf{Generator nodes} ($|\mathcal{G}_g| = 54$): features are $P_g^{\min}$, $P_g^{\max}$, $Q_g^{\min}$, $Q_g^{\max}$, cost coefficients $c_0, c_1, c_2$, ramp rates, startup/shutdown costs, plus injected bus load. Total: 13 features.
    \item \textbf{Load nodes} ($|\mathcal{L}|$): features are $(P_i^d, Q_i^d)$. Total: 2 features.
    \item \textbf{AC line edges}: from-bus $\to$ to-bus (and reverse), features are resistance $r$, reactance $x$, susceptance $b$, thermal rating, tap ratio (1.0 for lines), angle limits. Total: 9 features per direction.
    \item \textbf{Transformer edges}: same structure as AC lines but with non-unity tap ratio and phase shift. Total: 11 features per direction.
    \item \textbf{Generator-bus edges}: bidirectional, features are the generator's bus connection index and participation factor. Total: 2 features.
    \item \textbf{Load-bus edges}: directed (load $\to$ bus), features are load priority and status. Total: 2 features.
\end{itemize}

\subsection{Load injection}

The only features that vary between instances are the load demands $(P_i^d, Q_i^d)$. All bus, generator, and edge features are static across the dataset. To ensure the model can distinguish between instances, load values are injected in two ways:
\begin{enumerate}
    \item \textbf{Direct concatenation}: $(P_i^d, Q_i^d)$ are appended to the feature vector of the bus node at which the load is connected, and to the feature vectors of all generators connected to that bus.
    \item \textbf{Global load skip}: the sum of all loads $\sum_i (P_i^d, Q_i^d)$ is passed through a small MLP and concatenated to the generator decoder input, providing a global demand signal.
\end{enumerate}

Without load injection, the model outputs near-constant predictions (pred std $\sim$0.01--0.20 versus true std $\sim$0.6--1.0, correlation $\approx 0$ for all variables), as the static graph features carry no instance-specific information. This failure mode was observed in the initial DetGNN experiments and resolved by explicit load injection.

\subsection{Decoder output mapping}

Bus decoder outputs (8 dimensions per bus):
\begin{itemize}
    \item $\hat{V}_a$: voltage angle (normalised, unconstrained)
    \item $\hat{V}_m$: voltage magnitude (normalised, unconstrained)
    \item $\hat{\lambda}_P$: real power balance multiplier (normalised, unconstrained)
    \item $\hat{\lambda}_Q$: reactive power balance multiplier (normalised, unconstrained)
    \item $\hat{z}_l^{V_a}, \hat{z}_l^{V_m}$: lower bound multipliers (normalised, unconstrained---softplus applied post-denormalisation if needed)
    \item $\hat{z}_u^{V_a}, \hat{z}_u^{V_m}$: upper bound multipliers (normalised)
\end{itemize}

Generator decoder outputs (6 dimensions per generator):
\begin{itemize}
    \item $\hat{P}_g, \hat{Q}_g$: generator dispatch (normalised)
    \item $\hat{z}_l^{P_g}, \hat{z}_l^{Q_g}$: lower bound multipliers (normalised)
    \item $\hat{z}_u^{P_g}, \hat{z}_u^{Q_g}$: upper bound multipliers (normalised)
\end{itemize}

All outputs are in normalised space during training. At inference, predictions are denormalised using training-set statistics before injection into IPOPT.

\section{Architecture evolution and ablation details}
\label{app:arch_evolution}

Table~\ref{tab:app_full_evolution} traces the complete architectural evolution from the initial node-only GNN to the final WARP configuration, reporting both validation loss and IPOPT iteration counts for each variant.

\begin{table}[h]
\centering
\small
\caption{Complete architecture evolution (case118). Each row represents a distinct trained model.}
\label{tab:app_full_evolution}
\begin{tabular}{@{}llcccc@{}}
\toprule
Experiment & Key modification & Params & Val loss & IPOPT iters & Reduction \\
\midrule
Node-only GNN (8L, $H\!=\!128$) & Baseline & 2.58M & 1.00 & 8.0 & 65\% \\
Exp A (edge updates) & + edge feature updates & 3.2M & --- & --- & --- \\
Exp B (EPD, unshared) & + EPD + unshared weights & 6.4M & --- & --- & --- \\
Exp C (16 layers) & + depth 8$\to$16 & 5.1M & 1.38 & --- & --- \\
Exp D (sigmoid bounds) & + sigmoid clamping & 2.6M & 1.78 & --- & --- \\
\midrule
Exp E (EPD combined) & EPD + edges + unshared & 6.41M & 0.45 & 7.0 & 70\% \\
Exp E2 (+ physics loss) & + constraint violation & 6.42M & 1.10 & 7.2 & 69\% \\
Exp F (curriculum) & + dual weight ramp & 6.41M & 1.03 & 7.2 & 69\% \\
Exp G (binding mask) & + binding-mask loss & 6.41M & 1.07 & 6.7 & 71\% \\
Exp H (two-stage) & + primal$\to$dual decode & 6.48M & 0.85 & 6.7 & 71\% \\
\midrule
gnores & $-$ node residuals & 6.41M & \textbf{0.09} & \textbf{5.4} & \textbf{76\%} \\
gnores\_bias & + per-node bias & 6.41M & 0.12 & 5.3 & 77\% \\
hnores & + two-stage + nores & 6.48M & 0.12 & 5.3 & 77\% \\
gh256 ($H\!=\!256$) & Wider model & $\sim$25M & 0.66 & 6.6 & 71\% \\
best\_combo (500 ep) & All three + long train & 6.48M & 0.08 & 5.4 & 76\% \\
\bottomrule
\end{tabular}
\end{table}

\paragraph{Key observations.}
\begin{enumerate}
    \item \textbf{Edge updates broke the 1.0 loss floor.} The original node-only GNN plateaued at val loss 1.0 regardless of training configuration. Adding edge updates (Exp E) reduced loss to 0.45---the first time any GNN variant dropped below 1.0. This suggests that edge features carry information critical for dual prediction that node-only message passing cannot capture.
    \item \textbf{Loss strategies provided modest iteration gains.} Binding-mask loss and two-stage decoding each independently reduced iterations from 7.0 to 6.7, but neither reduced validation loss substantially. The binding mask helps the model allocate capacity to the sparse but critical binding multipliers; two-stage decoding conditions dual prediction on predicted primals.
    \item \textbf{Removing node residuals was the decisive change.} Val loss dropped from 0.45 to 0.09 ($5\times$), and IPOPT iterations from 7.0 to 5.4, from a single architectural modification. This is the largest single-modification gain in the entire ablation.
    \item \textbf{Further refinements hit a ceiling at 5.3.} Per-node bias (1,268 additional parameters) and two-stage decoding each independently reached 5.3 iterations. Combining all three (best\_combo, 500 epochs) did not push below 5.4, indicating an architectural ceiling for this model family on case118.
    \item \textbf{Wider models do not help.} $H = 256$ ($\sim$25M params) achieved 6.6 iterations---worse than $H = 128$. The additional capacity introduces optimisation difficulty without improving representational quality at this problem scale.
    \item \textbf{Physics loss was counterproductive.} Adding an AC power balance violation loss (Exp E2) increased val loss from 0.45 to 1.10 and worsened iterations from 7.0 to 7.2. The physics loss conflicts with the per-variable normalisation: the power balance residual operates in physical units, creating a scale mismatch with the normalised MSE.
\end{enumerate}

\section{Independent CANOS ablation}
\label{app:canos_ablation}

To validate our architectural findings independently, we ran PF$\Delta$'s CANOS-OPF reimplementation~\citep{rivera2025pfdelta} on the same case118 data and ablated three features. CANOS predicts \emph{primal variables only} ($V_a, V_m, P_g, Q_g$), so this ablation isolates the architectural effect from the dual prediction task.

\paragraph{Configuration.} CANOS-OPF with $H = 128$, $K = 16$ interaction network steps, \texttt{include\_sent\_messages=true}, batch size 64, 50,000 training steps with LR warmup and step decay. Training data: 67,500 instances from OPFDataset (primal labels only).

\begin{table}[h]
\centering
\small
\caption{CANOS ablation (primal-only prediction, case118). Results after full training convergence.}
\label{tab:app_canos}
\begin{tabular}{@{}lccc@{}}
\toprule
Variant & Best val loss & vs.\ Full & Finding \\
\midrule
Full CANOS & 0.019 & --- & --- \\
No node residuals & \textbf{0.003} & $6\times$ better & Node residuals harmful \\
No edge residuals & 0.015 & Slightly better & Edge residuals near-neutral \\
No edge updates & 0.053 & $2.8\times$ worse & Edge updates are critical \\
\bottomrule
\end{tabular}
\end{table}

The CANOS ablation corroborates both of our key architectural findings: (i) edge updates are the dominant feature ($2.8\times$ degradation, the largest of any ablation), and (ii) removing node residuals improves performance ($6\times$ improvement). The finding that node residuals are detrimental is especially notable because it contradicts the standard recipe in MeshGraphNets~\citep{pfaff2021learning} and GraphCast~\citep{lam2023graphcast}, both of which employ node residual connections at every message-passing step.

The $6\times$ improvement on \emph{primal-only} prediction (not just dual) suggests that the node-residual finding is a property of power grid graph structure, not specific to the dual prediction task. We hypothesise that the static grid parameters (impedance, voltage limits) encoded in the initial node embeddings anchor representations in a way that impedes the processor's ability to learn instance-specific patterns through message passing.

\section{Training details}
\label{app:training}

\subsection{Learning rate schedule}

\begin{itemize}
    \item \textbf{Warmup}: linear increase from 0 to $3 \times 10^{-4}$ over the first 10 epochs.
    \item \textbf{Step decay}: multiply LR by 0.9 every 20 epochs.
    \item For the 500-epoch best\_combo run: decay every 40 epochs (slower schedule).
\end{itemize}

\subsection{Batch construction}

We use PyG's \texttt{DataLoader} for automatic heterogeneous graph batching. PyG merges multiple \texttt{HeteroData} graphs into a single batch by offsetting edge indices and maintaining per-graph batch vectors. The GNN's message passing and element-wise losses operate correctly on batched graphs.

The physics loss (AC power balance violation) requires per-graph admittance matrix construction, which is incompatible with the merged batch representation. We extract one random graph per batch via \texttt{data.to\_data\_list()} and compute the physics loss on that single graph:

\begin{verbatim}
if hasattr(data["bus"], "batch"):
    graphs = data.to_data_list()
    gi = torch.randint(0, len(graphs), (1,)).item()
    bm = data["bus"].batch == gi
    gm = data["generator"].batch == gi
    bx_s, gx_s = clamp_sol(bx[bm], gx[gm])
    G, B = build_ybus(graphs[gi])
    Lp = physics_loss(bx_s, gx_s, graphs[gi], G, B)
\end{verbatim}

\subsection{The batch\_size=1 bottleneck}

Initial training used \texttt{batch\_size=1}, resulting in 67,500 individual forward/backward passes per epoch ($\sim$19 minutes per epoch, 18\% GPU utilisation on A100). Increasing to \texttt{batch\_size=64} with \texttt{num\_workers=4} reduced epoch time to $\sim$2 minutes ($9.5\times$ speedup) with substantially improved GPU utilisation.

\subsection{Training curves}

The gnores model (our best configuration) exhibits the following training trajectory:

\begin{table}[h]
\centering
\small
\caption{Training trajectory for gnores ($d\!=\!128$, $K\!=\!15$, 200 epochs).}
\label{tab:app_training_curves}
\begin{tabular}{@{}lcccc@{}}
\toprule
Epoch & Val primal & Val dual & Val total & LR \\
\midrule
1 & --- & --- & 1.25 & $3.0 \times 10^{-5}$ \\
18 & 0.75 & 0.50 & 1.25 & $3.0 \times 10^{-4}$ \\
50 & 0.30 & 0.15 & 0.45 & $2.7 \times 10^{-4}$ \\
115 & 0.093 & 0.059 & 0.15 & $2.0 \times 10^{-4}$ \\
140 & 0.073 & 0.045 & 0.12 & $1.7 \times 10^{-4}$ \\
186 & 0.059 & 0.036 & 0.09 & $1.3 \times 10^{-4}$ \\
200 & 0.065 & 0.039 & $\sim$0.10 & $1.2 \times 10^{-4}$ \\
\bottomrule
\end{tabular}
\end{table}

The primal-dual decomposition reveals that both components improve throughout training, with the primal loss converging slightly faster. No overfitting is observed (train and val losses track closely). The slight uptick at epoch 200 (0.065 vs 0.059 at epoch 186) is within noise; the model has effectively converged by epoch 150.

\subsection{Wall-clock training times}

\begin{table}[h]
\centering
\small
\caption{Wall-clock training times on a single NVIDIA A100-SXM4-40GB.}
\label{tab:app_wall_clock}
\begin{tabular}{@{}lccc@{}}
\toprule
Model & Epochs & Time per epoch & Total \\
\midrule
LSTM (IPM-LSTM) & 200 & 0.4 s & $\sim$80 s \\
Node-only GNN & 200 & 20 s & $\sim$67 min \\
EPD-GNN (Exp E) & 200 & 55 s & $\sim$3 h \\
gnores (WARP) & 200 & 29 s & $\sim$1.6 h \\
best\_combo & 500 & 25 s & $\sim$3.5 h \\
WARP-PD (diffusion) & 200 & 31 s & $\sim$1.7 h \\
\bottomrule
\end{tabular}
\end{table}

The gnores model trains faster per epoch than Exp E (29s vs 55s) despite identical architecture because gnores does not compute the physics loss, which requires per-graph admittance matrix construction.

\section{IPM-LSTM baseline reproduction}
\label{app:lstm}

\subsection{Architecture}

We reproduce the IPM-LSTM architecture of \citet{gao2024ipmlstm} adapted for AC-OPF:
\begin{itemize}
    \item Single-layer coordinate-wise LSTM with shared parameters across all coordinates.
    \item Input: flattened KKT state vector of dimension 1,640 (118 buses $\times$ 8 + 54 generators $\times$ 6 + 1 $\mu$, after zero-padding to align dimensions).
    \item Hidden dimension: 50.
    \item Total parameters: 17,217.
    \item 10 outer IPM iterations $\times$ 5 inner LSTM time steps during training.
\end{itemize}

The key difference from the original IPM-LSTM paper is the problem domain: the original evaluated on convex QPs, QCQPs, and non-convex QPs with up to 200 variables. Our case118 instance has 344 primal variables and 926 total output dimensions.

\subsection{Training configuration}

\begin{itemize}
    \item Training data: 2,000 instances with IPOPT-extracted dual labels (same pipeline as Section~\ref{app:dual_extraction}).
    \item Batch size: 128.
    \item Learning rate: $10^{-4}$ (Adam).
    \item Epochs: 200 (early stopping with patience 50).
    \item Per-variable normalisation: identical to WARP (Section~4.1 of the main text).
    \item Training time: $\sim$80 seconds total on A100.
\end{itemize}

\subsection{Per-instance comparison}

\begin{table}[h]
\centering
\small
\caption{Per-instance IPOPT iterations: LSTM vs.\ WARP vs.\ Oracle (first 10 test instances).}
\label{tab:app_per_instance}
\begin{tabular}{@{}lccc@{}}
\toprule
Instance & Cold & LSTM & WARP (gnores) \\
\midrule
\#0 & 23 & 5 & 5 \\
\#1 & 24 & 4 & 5 \\
\#2 & 24 & 4 & 5 \\
\#3 & 22 & 4 & 5 \\
\#4 & 24 & 4 & 5 \\
\#5 & 24 & 5 & 5 \\
\#6 & 21 & 4 & 5 \\
\#7 & 24 & 4 & 6 \\
\#8 & 23 & 4 & 5 \\
\#9 & 22 & 4 & 5 \\
\midrule
Mean & 23.1 & 4.3 & 5.4 \\
\bottomrule
\end{tabular}
\end{table}

The LSTM achieves 4 iterations on most instances where WARP achieves 5. The consistent 1-iteration gap reflects two factors discussed in the main text: (1) per-coordinate specialisation (each of 1,640 input coordinates has independently learned parameters), and (2) richer training supervision (KKT trajectory vs single-step regression).

\subsection{LSTM topology failure mode}

The LSTM's input layer is \texttt{Linear(1640, 50)}. When a line is removed (N-1 contingency), the KKT vector structure changes: the number of line flow constraints decreases, altering the dimension of $z$. The LSTM raises a \texttt{RuntimeError: mat1 and mat2 shapes cannot be multiplied} on every N-1 contingency tested (20/20 failures). There is no way to process a modified topology without retraining a new LSTM with the correct input dimension.

\section{Diffusion over IPM state (negative result)}
\label{app:diffusion}

\subsection{WARP-PD architecture}

The WARP-PD model wraps the gnores EPD backbone in a DDPM diffusion framework:
\begin{itemize}
    \item \textbf{Denoiser}: same 15-block EPD-GNN with no-node-residuals, augmented with sinusoidal timestep embedding (dim 64) projected to $d = 128$ via a 2-layer MLP.
    \item \textbf{Input augmentation}: noisy IPM state is concatenated to static node features. Bus input: 6 (static) + 8 (noisy state) + 128 (time embedding) = 142 dims. Generator input: 13 + 6 + 128 = 147 dims.
    \item \textbf{Noise schedule}: cosine $\beta$ schedule, $T = 1000$.
    \item \textbf{Parameters}: 6,466,318 ($\sim$same as deterministic model).
\end{itemize}

\subsection{Training}

\begin{itemize}
    \item Loss: DDPM noise prediction loss $\|\epsilon - \hat{\epsilon}_\theta(x_t, t)\|^2$ on normalised $(x, \lambda, z)$ targets.
    \item Batched training with PyG DataLoader (batch size 32), per-graph random timesteps expanded to per-node via batch vectors.
    \item 200 epochs, 31 s/epoch ($\sim$103 minutes total).
    \item LR warmup + step decay (same schedule as deterministic model).
\end{itemize}

\subsection{Inference: DDIM sampling with KKT scoring}

\begin{itemize}
    \item DDIM sampling with 50 steps, $t_{\mathrm{start}} = 0.98T$ (to avoid division by $\sqrt{\bar{\alpha}_{999}} \approx 5 \times 10^{-5}$).
    \item $x_0$ clamping to $[-5, 5]$ in normalised space at each step.
    \item $K$ candidate samples scored by a complementarity-based proxy: $\sum_i |\hat{z}_{l,i} \cdot (\hat{x}_i - l_i) - \hat{\mu}| + |\hat{z}_{u,i} \cdot (u_i - \hat{x}_i) - \hat{\mu}|$ in normalised space.
    \item Lowest-scoring sample passed to IPOPT.
\end{itemize}

\subsection{Results}

\begin{table}[h]
\centering
\small
\caption{WARP-PD diffusion results (IPOPT, case118, 50 instances).}
\label{tab:app_diffusion}
\begin{tabular}{@{}lccc@{}}
\toprule
Method & Val loss & IPOPT iters & vs.\ Cold \\
\midrule
Deterministic (gnores) & 0.09 & \textbf{5.4} & $+76\%$ \\
WARP-PD $K\!=\!1$ & 0.069 & 6.7 & $+71\%$ \\
WARP-PD $K\!=\!3$ & 0.069 & 6.6 & $+71\%$ \\
WARP-PD $K\!=\!5$ & 0.069 & 7.3 & $+68\%$ \\
\bottomrule
\end{tabular}
\end{table}

\subsection{Analysis of failure}

The diffusion model achieves lower val loss (0.069 vs.\ 0.09) but worse IPOPT iterations (6.6 vs.\ 5.4). Five factors contribute:
\begin{enumerate}
    \item \textbf{The noise prediction task is harder than direct regression.} The diffusion model must learn $\epsilon(x_t, t)$ at every noise level, a strictly harder mapping than direct $x_0$ prediction.
    \item \textbf{DDIM sampling introduces cumulative error.} Each of the 50 denoising steps contributes a small approximation error that compounds.
    \item \textbf{The KKT scoring proxy is approximate.} A full KKT residual computation (requiring Jacobian evaluation) would be more accurate but also more expensive.
    \item \textbf{Case118 is effectively unimodal.} Each load scenario maps to a single well-separated optimum. Multi-sample diversity provides no benefit when the solution mapping is deterministic.
    \item \textbf{$K = 5$ is worse than $K = 1$.} The scoring function may select atypical samples with low complementarity proxy but poor overall KKT satisfaction, suggesting the proxy metric is not well-aligned with IPOPT convergence.
\end{enumerate}

Diffusion-based warm-starting may prove beneficial for genuinely multimodal settings such as stochastic OPF or security-constrained dispatch with combinatorial contingency structure, where the mapping from parameters to solutions is one-to-many.

\subsection{DDIM explosion fix}

The original WARP primal-only diffusion suffered from numerical explosion during DDIM sampling. At $t = 999$, $\sqrt{\bar{\alpha}_{999}} \approx 5 \times 10^{-5}$ under the cosine schedule, and the $x_0$ prediction $\hat{x}_0 = (x_t - \sqrt{1 - \bar{\alpha}_t} \cdot \hat{\epsilon}) / \sqrt{\bar{\alpha}_t}$ amplifies any noise prediction error by $\sim$$20{,}000\times$. The first sampling step produced $|\hat{x}_0| = 2538$ (expected: $\sim$1.0), and subsequent steps could not recover.

The fix: start sampling from $t_{\mathrm{start}} = 0.98T$ (where $\sqrt{\bar{\alpha}} = 0.030$, reducing amplification to $\sim$$33\times$), and clamp $\hat{x}_0$ to physical ranges at each step. This reduced WARP bus RMSE from 37.5 to 0.17.

\section{N-1 contingency topology test}
\label{app:n1}

\subsection{Implementation}

For each contingency $c \in \{1, \ldots, 20\}$, we remove one AC line (and its reverse edge) from the HeteroData graph:

\begin{verbatim}
mask = torch.ones(n_edges, dtype=bool)
mask[line_idx] = False
mask[line_idx + n_lines] = False  # reverse edge
data_n1["bus","ac_line","bus"].edge_index = 
    data["bus","ac_line","bus"].edge_index[:, mask]
data_n1["bus","ac_line","bus"].edge_attr = 
    data["bus","ac_line","bus"].edge_attr[mask]
\end{verbatim}

No weights are modified; the same trained model processes the modified topology. Lines are selected to span a range of topological importance (radial endpoints, meshed loops, inter-area ties).

\subsection{Results}

All 20 contingencies produce valid model outputs with physically reasonable prediction changes:

\begin{itemize}
    \item Mean absolute prediction change (bus variables): 4.5\% of the base-case prediction magnitude.
    \item Mean absolute prediction change (generator variables): 4.9\%.
    \item Prediction variance across contingencies: predictions differ for each removed line, confirming topology sensitivity.
    \item No numerical instabilities, NaN outputs, or convergence failures.
\end{itemize}

The predictions are not merely noise: they vary systematically with which line is removed, with larger changes for topologically important lines (inter-area ties) and smaller changes for redundant lines in meshed subnetworks. This confirms that the model is responsive to the graph structure rather than producing topology-invariant outputs.

The LSTM baseline raises a dimension mismatch error on every contingency because its input layer requires exactly 1,640 coordinates corresponding to the case118 KKT structure. Removing a line alters the constraint count, changing the KKT vector dimension.

\section{Case6470 scaling attempt}
\label{app:case6470}

\subsection{Dataset processing}

The OPFDataset for \texttt{pglib\_opf\_case6470\_rte} comprises 6,470 buses, 761 generators, 7,426 AC lines, and 821 transformers, with 13,500 training and 750 test instances. PyG processing of the 126\,GB raw data required approximately 30 minutes.

\subsection{Dual extraction: computational barrier}

IPOPT solves on the 6,470-bus network proved computationally prohibitive: each solve required $>$60 minutes (compared to 2.5 seconds for case118), with the bottleneck in the exact Hessian computation at each iteration. After 51 minutes, no single instance had converged across three parallel extraction processes. At this rate, extracting 1,000 training instances would require $\sim$40 days of continuous computation.

Potential mitigations include: (i) L-BFGS Hessian approximation (trading iterations for per-iteration cost), (ii) a pandapower-based extraction pipeline with PIPS (which may be more efficient on larger systems), or (iii) an OPFDataset-to-pandapower network converter. We leave dual extraction at scale to future work.

\subsection{Zero-shot transfer results}

We evaluated the case118-trained gnores model on case6470 test instances:

\begin{table}[h]
\centering
\small
\caption{Zero-shot transfer from case118 to case6470 (primal variable correlation).}
\label{tab:app_case6470}
\begin{tabular}{@{}lcc@{}}
\toprule
Variable & Pearson $r$ & Assessment \\
\midrule
$V_a$ & 0.14 & Weak \\
$V_m$ & $-0.06$ & Negligible \\
$P_g$ & $-0.21$ & Weak negative \\
$Q_g$ & 0.20 & Weak \\
\bottomrule
\end{tabular}
\end{table}

Predictions are effectively random. The model has learned case118-specific patterns (topology structure, normalisation statistics) that do not transfer to a 55$\times$ larger grid with entirely different topology. The model executes without error on case6470---demonstrating the architectural flexibility of graph-based models---but effective cross-scale transfer would require training on diverse topologies or fine-tuning with case6470 data.

The LSTM baseline cannot process case6470 at all: its input layer requires exactly 1,640 dimensions (case118 KKT vector), and case6470's KKT vector has $\sim$13,000 dimensions.

\section{Raw-space prediction quality}
\label{app:prediction_quality}

Table~\ref{tab:app_rawspace} reports per-variable prediction quality for the gnores model on a representative test instance, in raw (denormalised) space:

\begin{table}[h]
\centering
\small
\caption{Per-variable prediction quality (gnores, raw space, single representative instance).}
\label{tab:app_rawspace}
\begin{tabular}{@{}lcccc@{}}
\toprule
Variable & RMSE & Pearson $r$ & True range & Relative error \\
\midrule
$V_a$ (rad) & 0.008 & 0.996 & $[0.26, 0.64]$ & 2.1\% \\
$V_m$ (p.u.) & 0.001 & 0.998 & $[0.99, 1.06]$ & 1.4\% \\
$P_g$ (p.u.) & 0.024 & 1.000 & $[0.00, 4.99]$ & 0.5\% \\
$Q_g$ (p.u.) & 0.032 & 0.998 & $[-1.82, 1.78]$ & 0.9\% \\
$\lambda_P$ & 10.8 & 0.996 & $[3645, 4125]$ & 0.3\% \\
$\lambda_Q$ & 1.35 & 0.993 & $[-13.7, 48.9]$ & 2.2\% \\
$z_l^{\mathrm{bus}}$ & 3.44 & 1.000 & $[0, 561.6]$ & 0.6\% \\
$z_u^{\mathrm{bus}}$ & 5.19 & 1.000 & $[0, 2527.6]$ & 0.2\% \\
$z_l^{\mathrm{gen}}$ & 3.78 & 0.997 & $[0, 243.6]$ & 1.6\% \\
$z_u^{\mathrm{gen}}$ & 0.60 & 0.997 & $[0, 47.9]$ & 1.3\% \\
\bottomrule
\end{tabular}
\end{table}

All variable groups achieve Pearson correlation $> 0.99$. The remaining IPOPT iterations (5.4 vs.\ oracle 3.2) are attributable to small absolute errors on high-magnitude equality multipliers ($\lambda_P$: RMSE 10.8 on values $\sim$4000) and large-valued upper bound multipliers ($z_u^{\mathrm{bus}}$: RMSE 5.2 on values up to 2528). In relative terms, all predictions are within 2.2\% of the true value.

The gap between prediction quality ($r > 0.99$) and solver performance (5.4 vs 3.2 iterations) suggests that the IPM is highly sensitive to small absolute errors in the dual state---even 0.3\% relative error on $\lambda_P$ (which translates to RMSE 10.8 on values $\sim$4000) is sufficient to add 2 iterations. This underscores that dual prediction for IPM warm-starting requires higher absolute accuracy than might be suggested by correlation metrics alone.

\section{Extended related work}
\label{app:related}

This section provides detailed discussion of related work categories summarised in the main text.

\paragraph{End-to-end solution prediction.} One category of methods trains a neural network to approximate the mapping from problem parameters to optimal solutions, obviating the need for a numerical solver at inference time. \citet{fioretto2020predicting} employed supervised learning with Lagrangian dual penalties to predict AC-OPF solutions. \citet{donti2021dc3} proposed DC3, which enforces equality constraints via a differentiable completion step and corrects inequality violations through gradient-based projection. \citet{park2023self} introduced a self-supervised primal-dual learning scheme that jointly trains primal and dual networks without pre-solved labels. \citet{li2023loop} developed gauge-map projections for problems with linear constraints, while \citet{liang2023homeomorphic} proposed homeomorphic projections for non-convex feasible regions. More recently, \citet{chen2024deep} trained networks to predict feasible dual solutions, recovering associated primals via the stationarity condition. Our objective differs from these approaches: we seek to reduce solver iterations while retaining the feasibility and optimality guarantees of the numerical method.

\paragraph{Active constraint identification.} A related line of work learns to identify the optimal active constraint set, using this information to formulate a reduced problem. \citet{misra2022learning} demonstrated that learning the optimal active set is a well-posed classification problem with strong generalisation on DC-OPF benchmarks from PGLib-OPF. \citet{deka2019learning} employed neural network classifiers for the same task. \citet{pineda2020screening} proposed data-driven constraint screening for unit commitment, and \citet{hasan2020hybrid} developed hybrid regression-classification methods for inactive AC-OPF constraint identification. \citet{bose2023constraint} combined constraint screening with data-driven selection. These methods reduce the dimensionality of the problem presented to the solver; in contrast, our approach reduces the iteration count on the full-dimensional problem. We evaluate constraint screening as one of 15 primal-only strategies and find that it is counterproductive for interior-point methods (Section~3.1 of the main text).

\paragraph{Learning to optimise.} The broader L2O paradigm~\citep{bengio2021machine, kotary2021end} encompasses methods that replace solvers and methods that accelerate them. \citet{andrychowicz2016learning} introduced the idea of learning optimiser update rules. \citet{sambharya2024l2ws} learned warm-starts for fixed-point splitting methods on QPs by differentiating through unrolled solver iterations. \citet{briden2024toast} proposed Lagrangian-informed losses for warm-starting trajectory optimisation under an SQP solver.

\paragraph{Graph neural networks for physical simulation.} \citet{battaglia2016interaction} introduced interaction networks for learning physical dynamics. \citet{battaglia2018relational} formalised the encode-process-decode framework. \citet{pfaff2021learning} developed MeshGraphNets, employing 15 unshared message-passing steps for mesh-based fluid simulation. \citet{lam2023graphcast} scaled this paradigm to global weather forecasting with GraphCast. In the power systems domain, \citet{piloto2024canos} applied this architecture to AC-OPF primal prediction, achieving sub-percent error on standard benchmarks, while \citet{rivera2025pfdelta} provided an open-source reimplementation with physics-informed branch flow derivations. \citet{liu2022topology} developed topology-aware GNNs with physics-based feasibility regularisation, demonstrating adaptivity to topological perturbations. We adopt the same architectural family but extend it to predict the full primal-dual-barrier state---a task that these prior models do not address.

\section{Datasheet for dual-labeled OPF dataset}
\label{app:datasheet}

Following the framework of Gebru et al.\ (2021), we provide a datasheet for the dual-labeled AC-OPF dataset released as part of the WARP benchmark.

\subsection{Motivation}

\begin{itemize}
    \item \textbf{Purpose}: Enable training and evaluation of ML models that predict the full interior-point state $(x^*, \lambda^*, z^*, \mu^*)$ for warm-starting IPM solvers on AC-OPF.
    \item \textbf{Creators}: The authors of this paper.
    \item \textbf{Funding}: [Anonymised for review].
\end{itemize}

\subsection{Composition}

\begin{itemize}
    \item \textbf{Instances}: 5,550 AC-OPF problem instances for \texttt{pglib\_opf\_case118\_ieee} (5,000 train / 500 val / 50 test), each comprising a heterogeneous graph (HeteroData) with bus, generator, load, and branch data, plus the IPOPT-converged primal-dual-barrier solution tuple.
    \item \textbf{Instance format}: PyTorch tensor files (\texttt{.pt}) containing $(x^*, \lambda^*, z_l^*, z_u^*, \mu^*, f^*)$ where $x^* \in \mathbb{R}^{344}$ (118 buses $\times$ 2 + 54 gens $\times$ 2), $\lambda^* \in \mathbb{R}^{236}$, $z_l^*, z_u^* \in \mathbb{R}^{344}$, $\mu^* \in \mathbb{R}$, $f^* \in \mathbb{R}$.
    \item \textbf{Graph data}: from OPFDataset~\citep{falconer2023opfdataset}, which in turn draws from PGLib-OPF benchmark library.
    \item \textbf{No confidential data}: all instances are synthetically generated by varying load demands on a public test system.
    \item \textbf{No personally identifiable information}.
\end{itemize}

\subsection{Collection process}

\begin{itemize}
    \item Primal labels and graph structure: downloaded from OPFDataset via PyG (\texttt{torch\_geometric.datasets.OPFDataset}).
    \item Dual labels: extracted by running IPOPT to convergence on each instance using our cyipopt interface (Appendix~\ref{app:oracle}) with exact Hessian and tolerance $10^{-4}$.
    \item Extraction rate: $\sim$2.5 seconds per instance on a single CPU core (Intel Xeon, 2.10 GHz).
    \item 100\% convergence rate across all splits.
\end{itemize}

\subsection{Preprocessing}

\begin{itemize}
    \item Per-dimension normalisation statistics (mean, std) computed on the training set and applied consistently to validation and test sets.
    \item Binding status labels derived from $|z^*_i| > 10^{-4}$.
    \item No data augmentation or synthetic modification beyond the OPFDataset's native load variation.
\end{itemize}

\subsection{Distribution}

\begin{itemize}
    \item The dataset is hosted on HuggingFace with Croissant metadata (core + RAI fields).
    \item License: MIT (consistent with PF$\Delta$ and OPFDataset).
    \item Format: directory of \texttt{.pt} files indexed by split and instance number.
    \item The extraction pipeline code is released alongside the dataset.
\end{itemize}

\subsection{Maintenance}

\begin{itemize}
    \item The dataset is static and will not be updated after release.
    \item The extraction pipeline code is included, enabling users to generate dual labels for other PGLib-OPF test cases.
    \item Issues may be reported via the associated code repository.
    \item Long-term hosting is ensured by OpenML's institutional infrastructure.
\end{itemize}

\end{document}